\documentclass{ieeeaccess}

\usepackage{amsmath}
\usepackage{color}
\usepackage{array}
\usepackage{stfloats}
\usepackage{amsthm} 
\usepackage{amssymb}
\usepackage{float}
\usepackage[normalem]{ulem}
\usepackage{soul}
\usepackage{nccmath}
\usepackage{graphicx}
\usepackage{setspace}
\usepackage{booktabs}
\usepackage{subcaption}
\usepackage{mwe}
\usepackage{cite}
\usepackage{amsmath,amssymb,amsfonts}
\usepackage{textcomp}
\usepackage{xcolor}
\usepackage[utf8]{inputenc} 
\usepackage{kotex} 

\usepackage{algorithm}
\usepackage[noend]{algpseudocode}
\usepackage{algorithmicx}
\usepackage{grffile}
\usepackage{multirow}
\usepackage{mathtools}

\usepackage{xspace}
\def\BibTeX{{\rm B\kern-.05em{\sc i\kern-.025em b}\kern-.08em
    T\kern-.1667em\lower.7ex\hbox{E}\kern-.125emX}}

\begin{document}

\history{Date of publication xxxx 00, 0000, date of current version xxxx 00, 0000.}
\doi{10.1109/ACCESS.2017.DOI}

\title{Spatio-Temporal Split Learning for Privacy-Preserving Medical Platforms: Case Studies with COVID-19 CT, X-Ray, and Cholesterol Data}
\author{\uppercase{Yoo Jeong Ha}\authorrefmark{1}, 
\uppercase{Minjae Yoo}\authorrefmark{1},
\uppercase{Gusang Lee}\authorrefmark{1},
\uppercase{Soyi Jung}\authorrefmark{2},
\uppercase{Sae Won Choi}\authorrefmark{3},
\uppercase{Joongheon Kim}\authorrefmark{1} \IEEEmembership{Senior Member, IEEE}, and \uppercase{Seehwan Yoo}\authorrefmark{4} \IEEEmembership{Senior Member, IEEE}}
\address[1]{Department of Electrical and Computer Engineering, Korea University, Seoul, Korea (e-mails: annaha17@korea.ac.kr,
mj7015@korea.ac.kr, rntkd0917@korea.ac.kr, joongheon@korea.ac.kr)}
\address[2]{School of Software, Hallym University, Chuncheon, Korea (e-mail: jungsoyi20@gmail.com)}
\address[3]{Office of Hospital Information, Seoul National University Hospital, Seoul, Korea (e-mail: swc1@snu.ac.kr)}
\address[4]{Department of Mobile Systems Engineering, Dankook University, Yongin, Korea (e-mail: seehwan.yoo@dankook.ac.kr)}
\tfootnote{This work was supported by the Ministry of Health and Welfare (MHW), Korea (HI19C0572).}

\markboth
{Y.J. Ha \headeretal: Spatio-Temporal Split Learning for Privacy-Preserving Medical Platforms}
{Y.J. Ha \headeretal: Spatio-Temporal Split Learning for Privacy-Preserving Medical Platforms}

\corresp{Corresponding authors: Soyi Jung, Joongheon Kim, and Seehwan Yoo (e-mails: jungsoyi@korea.ac.kr, joongheon@korea.ac.kr, seehwan.yoo@dankook.ac.kr).}


\begin{abstract}
Machine learning requires a large volume of sample data, especially when it is used in high-accuracy medical applications. However, patient records are one of the most sensitive private information that is not usually shared among institutes. This paper presents spatio-temporal split learning, a distributed deep neural network framework, which is a turning point in allowing collaboration among privacy-sensitive organizations. Our spatio-temporal split learning presents how distributed machine learning can be efficiently conducted with minimal privacy concerns. The proposed split learning consists of a number of clients and a centralized server. Each client has only has one hidden layer, which acts as the privacy-preserving layer, and the centralized server comprises the other hidden layers and the output layer. Since the centralized server does not need to access the training data and trains the deep neural network with parameters received from the privacy-preserving layer, privacy of original data is guaranteed. We have coined the term, spatio-temporal split learning, as multiple clients are spatially distributed to cover diverse datasets from different participants, and we can temporally split the learning process, detaching the privacy preserving layer from the rest of the learning process to minimize privacy breaches. This paper shows how we can analyze the medical data whilst ensuring privacy using our proposed multi-site spatio-temporal split learning algorithm on Coronavirus Disease-19 (COVID-19) chest Computed Tomography (CT) scans, MUsculoskeletal RAdiographs (MURA) X-ray images, and cholesterol levels. 
\end{abstract}


\titlepgskip=-15pt

\maketitle
\section{Introduction}\label{sec:intro} 
\PARstart{}{} 

Machine learning in the medical field has transformed the way hospitals operate; diagnosis, surgical methods, and treatment plans have been devised in a much shorter time frame. Deep learning has been extensively used to identify, classify and quantify patterns in medical images. CT, magnetic resonance imaging (MRI), positron emission tomography (PET), X-ray, and ultrasound scans are a few examples of medical image techniques that have been used in the past several decades for early detection and treatment {\cite{medicine}}. To develop more algorithms in order to treat more patients, information is crucial in the medical sector. Hospitals, research facilities, pharmaceutical companies must possess an abundant amount of medical data to advance healthcare. Yet, when it comes to exposing medical data that include demographic information, consultation notes, immunizations, allergies, surgical history on patients, people are reluctant to provide private information. As more information is revealed on online platforms, people are stepping in to secure their privacy, and rightfully so. 

Medical records, perhaps the most private information on an individual, should be protected at all costs, yet it is not so easy to do with hospitals deploying deep neural networks to train algorithms using real patient data. Most hospitals maintain digital versions of patient charts like the electronic medical record (EMR). EMR is classified as highly sensitive data because it contains a patient’s medical and treatment history from an institute \cite{emr}. EMR data cannot be exposed to any other organization outside the hospital by regulation and is not shared with other healthcare providers unless the patient changes one’s primary physician. Yet, these medical records such as CT images or X-ray scans are crucial information for laboratories, specialists, clinicians, and even corporations beyond the health organization for research purposes. The personal information that tags along these medical data make it difficult for deep learning to develop. Here is where our proposed method comes in. Spatio-temporal split learning allows models to be trained without revealing the original raw data, which in our case are sensitive medical records. 

In split learning, the deep neural network is separated amongst the clients or end-systems and a centralized server~\cite{9265295,bakator2018,gupta2018,dsn19jeon,ictc19jeon}. 
The local end-system learns the model only up to a specific layer, typically the first hidden layer. 
The parameter updated from the first hidden layer is transferred to the centralized server where the rest of the computation is conducted. In this paper, this type of network structure that divides a deep neural network between one client and one server is coined the term single-client split learning to differentiate it with our proposed approach. 

Spatio-temporal split learning, the method devised in this paper, divides the deep neural network (temporal) among multiple clients allocated in geographically different locations (spatio) and one centralized server~\cite{dsn2020}. 
Each client has a privacy preserving layer, and begins learning process with patient's medical record.
Only after the privacy preserving layer, the model parameters are transferred to the central server to finish the training. 
Since training data is not shared among the clients, the privacy attack contingency is efficiently reduced. 
Note that the training data of medical applications are patients' medical record, which is one of the most sensitive private information.
In a realistic setting where numerous medical institutes collaborate to build a universal and high-accuracy deep neural network model, we can extend the proposed split learning to comprise more clients in different hospitals. Although the split learning structure mentioned in~\cite{dsn2020} is similar to our proposed method, our work explores the feasibility of applying spatio-temporal split learning to medical data to protect the privacy of patient's personal information. Our paper emphasizes the first setting where split learning is used in the medical field.

This paper explores the specific way in which multiple hospitals utilize the method of spatio-temporal split learning to collaboratively create a deep neural network model. The main contributions of this work are summarized below: 
\begin{itemize}
    \item This paper proposes an innovative deep learning approach to preserve the privacy of personal health data through split learning. 
    \item The proposed method resolves the issue of data-imbalance and overfitting, a problem that arises when a model is trained with insignificant amounts of data. With spatio-temporal learning, it is practically impossible for overfitting to occur since multiple clients with various amounts of data all contribute in training the one model.
    \item The learning method presented in this paper proves to be versatile. The proposed split learning method works with both numerical and image data for the prediction model and CNN model, respectively. 
    \item The performance improvements and novelty of our proposed approach are evaluated with various real-world medical data such as cholesterol data provided by Seoul National University Hospital (SNUH)\footnote{This study was approved by the Institutional Review Board of Seoul National University Hospital (No. C-1712-009-903) with a waiver of informed consent. No personally identifiable data was included in the dataset. Data used in this study was retrieved from Seoul National University Hospital's Common Data Model (CDM) database.}, COVID-19 chest CT scans, and MURA datasets.
\end{itemize}
The rest of this paper is organized as follows. Related works on distributed learning is presented in Section~\ref{sec:related}. Section~\ref{sec:3} presents the problem statement and the system model with design considerations. The experimental results are presented in Section~\ref{sec:4}. Finally, the paper concludes with Section~\ref{sec:5}. 

\section{Related Work}\label{sec:related}
Collaborative machine learning using medical data is limited due to regulations that prohibit patient information from leaving the hospital. Federated learning (FL), a basic form of distributed learning, is a commonly used technique that learns a deep neural network while maintaining the security of original data \cite{pieee21park,iotj2020kwon}. Similar to split learning, FL consists of several clients and a centralized server. Each client, such as a mobile device or a base station, has its own deep neural network that trains with its local data. Once the client learns the model, the final parameter from every client is encrypted and transmitted to the central server. Since only the local parameters are shared by clients, instead of training data, the privacy of the original training data is preserved. However, this can be a double-edged-sword. As individually trained models are integrated by averaging the parameter updates without the actual characteristics of original data, a drop in the accuracy of the models is inevitable. 

Nevertheless, the learning process of FL is very different from split learning. Both split learning and FL consists of a client and a centralized server. Split learning takes it one step further and literally splits the deep neural network into two sections. The input layer and first hidden layer reside with the client and the remaining hidden layers are located up in the centralized server. Thus, it is the server that holds and trains the one deep neural network. On the other hand, in FL, each client possesses a full deep learning model using its own training data. The individual clients have their local training dataset that does not leave its boundary. Only the updates made to the current global model from each client are sent to the server~\cite{yang2019federated}. Hence, only the encrypted individual updates from each client are exposed to the server. It is important to understand that FL has the entire deep learning model contained in each client; whereas, in split learning, the client only runs the first hidden layer of the deep neural network.  For FL, the server only holds the updates for each model ephemerally; while for split learning, the majority of the deep learning model is placed in the server.

\begin{figure*}
    \begin{center}
        \includegraphics[width=0.65\linewidth]{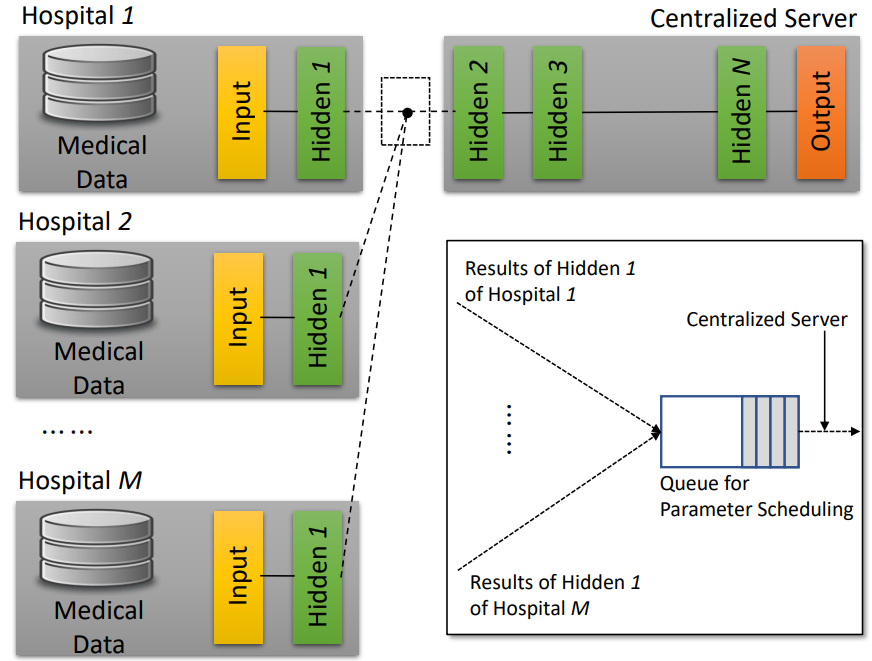}
    \end{center}
    \caption{ The overall architecture of training through our proposed spatio-temporal split learning. }
    \label{architecture}
\end{figure*}

One of the benefits of spatio-temporal split learning is its ability to prevent overfitting. Our proposed method is able to gather multiple clients to all collaborate in developing one big deep neural network. Therefore, even if a certain client does not hold an abundance of data, it can still benefit from the other clients that all contribute to training the model. Since countless data are gathered to train a deep neural network, overfitting will not occur in this novel spatio-temporal split learning procedure. 

Our model overcomes the problem of data-imbalance amongst the participating hospitals as well as overfitting. For example, if one medical facility possesses a minimum amount of data on MRI scans of brain tumors, due to this data-imbalance, the model is bound to be overfitted. Yet, if this particular medical facility participates in the spatio-temporal learning approach, it can collaborate with another institute that holds an ample number of scans of a brain tumor, thus, avoiding the problem of overfitting.

Moreover, this multi-client split learning allows different kinds of clients to participate in the learning of the deep neural network.
This innovative spatio-temporal split learning is applicable to various types of data. Image classification models and regression models can all be learned using the spatio-temporal split learning approach. The experiments conducted in this paper are done using two groups of data—numerical data and image data. COVID-19 chest CT scans and MURA datasets, which are both image datasets, and cholesterol data, a numerical dataset, are used to evaluate the performance of spatio-temporal split learning compared to single-site split learning. Our work emphasizes its prominence in the medical field where thousands of patients are waiting to be treated every day and require swift and accurate analysis of test scans.

Another major benefit of spatio-temporal split learning that is consistently highlighted in this paper is the preservation of privacy. Due to the learning process of split learning, where the original training does not get exposed to external networks, protection of personal information is guaranteed. Only the encrypted, highly distorted feature maps from every participating client are communicated to the external server. There are model inversion attack techniques~\cite{fredrikson2015model, zhang2020secret} that can abuse the neural network models to infer information about its training data. This powerful tool can raise valid concerns, yet, our spatio-temporal split learning is not affected by these model inversion attacks. This is because the current attacks are assumed on a fully intact model, and there is lacking research on the feasibility of such attacks on split learning models. Furthermore, the surface that can be attacked in a split learning environment is quite limited.

\section{Multi-Site Split Learning for spatially-Distributed Medical Platforms}\label{sec:3}

\subsection{System model}

The proposed spatio-temporal split learning revolutionizes the machine learning process, where several medical organizations or institutes collaborate with each other to reach a common goal—to establish a competent medical AI application that can assist in accurate patient treatment. 
The idea of multi-site split learning allows multiple clients to co-operate with each other in the most advantageous way. 
Spatio-temporal split learning allows clients to participate in machine learning studies, and contributes to build more accurate medical models without directly obtaining copious amounts of raw medical data. 
More importantly the proposed spatio-temporal split learning preserves patients privacy when we utilize invaluable medical information of another client organization including hospital, research institute, pharmaceutical companies.

The overall architecture of our proposed system is depicted in Fig.~\ref{architecture}. In the figure, the learning layers are split in two parts: the hospitals (i.e., clients) and centralized server. 
In our environment, hospitals are the clients that make requests to build a deep neural network model with its patient data. The hospital holds the input layer and the first hidden layer, and the centralized server has the rest of the layers; thus most of the computation occurs at the central server, where it builds up a high-accuracy medical neural network. Each hospital runs the training process only up to the first hidden layer. The feature map that is produced after going through the first hidden layer in each hospital is sent to the centralized server. The server has a queue for taking feature maps from different clients, allowing multiple clients to work asynchronously.

%
%

A hidden layer comprises of the convolution (Conv2D) and/or max-pooling (MaxPooling2D). When CT, MRI, or PET image data pass through the hidden layer, some information is lost due to the nature of convolutional and max-pooling operations. This is an essential operation in order to preserve privacy because the operations make it hard to distinguish a specific data item. Specifically, max pooling, and activation functions apply non-linear and non-reversible operation to the original data; thereby, once applied, it is hard to infer the original values from the feature map. As these medical images pass through more hidden layers in the centralized server, the images become even harder to identify a specific data element--hence, the privacy of the original data is strongly preserved. 

Besides the advantages in privacy, the proposed split learning makes the clients much light-weight because a client conducts computations only up to the first hidden layer. Computation-intensive layers are integrated into the central server, and the server can efficiently schedule the learning process based on the server's parameter queue.

\subsection{Algorithm Design and Pseudo-Code} 
\subsubsection{Algorithm design} 

Our spatio-temporal split learning consists of the client algorithm and the server algorithm. 
\begin{itemize}
    \item \textbf{Client algorithm:} The client preserves the privacy of data by 1) not exposing the raw data or 2) only disclosing the encrypted feature maps. The algorithm enforces the two properties. First, a client transfers the feature map only after the privacy-preserving layer is passed. That is, raw data is isolated from the machine learning pipeline. Second, the client algorithm adds enough noise to the image that it becomes difficult to infer the original data with only the information about the feature map.
    
    \item \textbf{Server algorithm:} The server gathers the outputs of multiple clients and then computes the remaining machine learning pipeline. The outputs are the feature maps after passing through the one hidden layer that practically encrypts the original data. These feature maps are trained along the multiple hidden layers positioned in the server to produce a deep neural network model. The only information the server receives is the encrypted feature map from the client. Since the raw data is not exposed to the server, privacy of personal information is guaranteed.
\end{itemize}

In addition, our algorithm introduces a parameter queue for receiving feature map from clients, as shown in Fig.~\ref{architecture}. The design has several benefits when we consider multiple clients. First, the server does not stop processing for incoming client data. While the server learns the model by calculating the parameters, it can receive data from another client. Second, the server can control the amount of input data from different clients. The amount of input data differs from clients because different hospitals have different number of patients, and different cases. 

The formal description of our proposed spatio-temporal split learning can be presented in a form of a pseudo-code, as shown in the following Sec.~\ref{sec:pc}.

\begin{algorithm}[t!]
\begin{algorithmic}[1]
\Statex $\hspace{-1.5em}$ \textbf{Require:}~ Batch size $B$, clients $\emph{C}$, number of clients $\emph{n}$, number of epoch $\emph{E}$, learning rate $\alpha$, target value ${y}$, predicted value ${\hat{y}}$, input data $I$, number of input data ${I_n}$, number of label ${l_a}$, and output convolution layer $O^{l}$.
\vspace{0.4em}

\Procedure{Client}{}\\
\textbf{For} {Client = \{1, $\dotsm$, $n$\}} \textbf{do}\\
	$\hspace{1.5em}$ \textbf{For} \textit{Training data set} = \{1, $\dotsm$, $x$\} \textbf{do}\\
	$\hspace{3em}$ Calculate Conv. \Comment{Eq.~\eqref{eq:Conv}}\\
	$\hspace{3em}$ $\triangleright$ $f_{c}$ = $\mathrm{Conv}(O^{l-1},w^{l},I_n,l_a) =net_{I_n,l_a}^{l}$\\
    $\hspace{1.5em}$ $\triangleright$ Send feature $f_{c}$ to server.\\
    $\hspace{1.5em}$  \textbf{End For}\\
 \textbf{End For}\\
\EndProcedure

\Procedure{Server}{}\\
    Receive input data from client : $f_{c}$\\
    Concatenate all features $\sum_{k=1}^{n}f_{c}^{k}$\\
\textbf{For} \textit{epoch = 1, $E$} \textbf{do}\\
    $\hspace{1.5em}$ \textbf{For} \textit{Training data set} \textbf{do}\\
        $\hspace{3em}$ Calculate Conv, and Pool \Comment{Eq.~\eqref{eq:Conv}, Eq.~\eqref{eq:Pool}}$\vspace{0.3em}$ \\
    	$\hspace{3em}$ $\triangleright$ $f_{c} = \mathrm{Conv}(O^{l-1},w^{l},I_n,l_a) = net_{I_n,l_a}^{l}$\\
    	$\hspace{3em}$ $\triangleright$ $f_{p}=\mathrm{Pool}(f_{c},I_m,I_a)$\\
    	$\hspace{3em}$ $\triangleright$ ${\hat{y}}$ is calculated using $I$, \eqref{eq:Conv}, and \eqref{eq:Pool}.\\
    $\hspace{1.5em}$ $\triangleright$ Calculate loss. \Comment{Eq.~\eqref{eq:msle}}\\
    $\hspace{1.5em}$ $\triangleright$ Update the model: update weights $w$ $\leftarrow$ $w$ $\cdot $ $\alpha$.\\
	    $\hspace{1.5em}$  \textbf{End For}\\
 \textbf{End For}
 \EndProcedure
 	\caption{Multi-client spatio-temporal split learning}
	\label{algorithm1}
	\end{algorithmic}
\end{algorithm}

\subsubsection{Pseudo-code}\label{sec:pc}

The pseudo-code of the proposed spatio-temporal split learning is presented in Algorithm 1. From (line 1) to (line 8), the split learning algorithm of a client is explained. A client runs only one hidden layer calculated by Equation~\ref{eq:Conv} in (line 4) ${\sim}$ (line 5). In (line 6), the calculated the parameter $f_{c}$ is sent to the server. In (line 10) to (line 22), the split learning algorithm in the server side is presented. 
The server basically runs the machine learning process. The only difference is that the server begins with the received feature set, that is placed on a parameter queue.
The computation for Conv2D and MaxPooling2D layers are formulated as following \eqref{eq:Conv} and \eqref{eq:Pool}, respectively, as depicted in (line 11) and (line 12), i.e.,
\begin{equation}
net_{mn}^{l} =\sum\limits_{i=0}^{size^{l}-1} \sum\limits_{j=0}^{size^{l}-1} (O_{m+i,n+j}^{l-1} \dots w_{i,j}^{l}+b^{l}),\label{eq:Conv}
\end{equation}
where $O^{l}$ is the output convolution layer, $w$ is the weight parameter, $b$ is the bias, $l$ is the layer, $i$ and $j$ are the row and column of the matrix, and size represents the number of neurons, and 
\begin{multline}
\mathrm{Pool}(x,i,j)= \\ \cfrac{\sum\limits_{m=1}^{size^{l}} \sum\limits_{m=1}^{size^{l}} x_{size^{l} \times {(i-1)} + m, size^{l} \times (j-1) + n }^{l-1}}{size^{l} \times size^{l}}, \label{eq:Pool}
\end{multline}
where $x$ is the matrix that was convoluted and serves as the input in the pooling layer, $l$ is the layer, $i$ and $j$ are the row and column of the matrix, and size represents the number of neurons.

\section{Experiments}\label{sec:4}

\subsection{Non-medical data}\label{sec:Data}

Proposed split learning works not only with medical images data, but also with general data. 
A recent study on split learning conducted in~\cite{dsn2020} shows the accuracy rate at each stage of the hidden layer, proving the optimal performance and maintaining data privacy. 
This experiment used CNN with cifar10  and a very similar setup to our experimental design. 
They paired one Conv2D layer and one MaxPooling 2D layer to comprise one hidden layer. 16, 32, 64, 128, and 256 filters were used for each of the five hidden layers with size 32x32. 

\begin{figure}[t!]
\centering
\setlength{\tabcolsep}{2pt}
\renewcommand{\arraystretch}{0.2}
\begin{tabular}{p{0.31\linewidth}p{0.02\linewidth}p{0.31\linewidth}p{0.02\linewidth}p{0.31\linewidth}}
\tabularnewline
\tabularnewline
\includegraphics[page=1, width=0.95\linewidth]{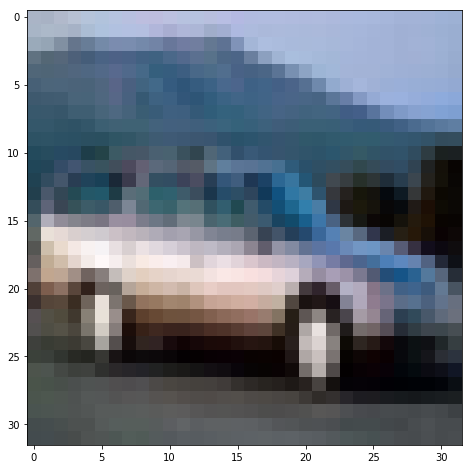} & {} &
\includegraphics[page=1, width=0.95\linewidth]{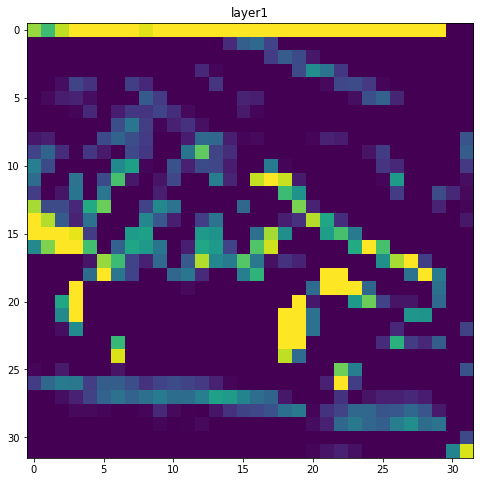} & {} &
\includegraphics[page=1, width=0.95\linewidth]{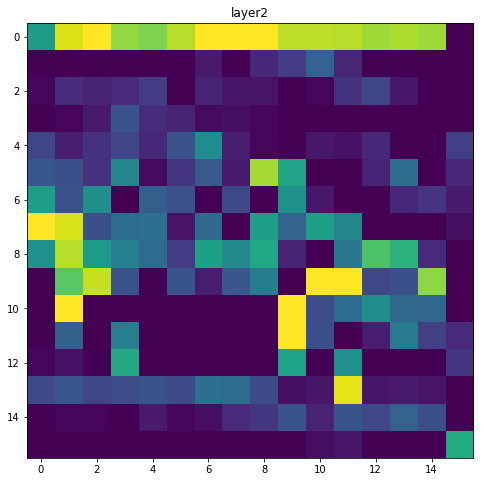}
\\
\tabularnewline
\tabularnewline
\centering(a) Original & {} &
\centering(b) Conv2D in $L_{1}$ & {} &
\centering(c) $L_{1}$
\tabularnewline
\end{tabular}
\caption{Image capture during deep neural network computation.} 
\label{fig:privacy}
\end{figure}

As summarized in Table~\ref{accuracylayer}, a classification accuracy of 71.09\% can be achieved when all of the five layers are located at the centralized server and computed within the server. The classification performance drops to 68.18\%, a fall of 2.91\%, when the first hidden layer is placed at the client's side and the rest at the server. There is only a trivial reduction in classification accuracy, yet this can be overlooked since the advantage of data privacy in hospital settings is much more significant. Paper~\cite{dsn2020} considered the extreme case where the first four layers of the neural network are computed by the client. That scenario had a performance of 65.66\%, which is only a 5.43\% degradation in accuracy. Thus it shows that a very minor fall in performance is sacrificed for a greater benefit in preserving privacy for medical data. By sacrificing a drop in performance of around 5\%, the privacy of the trained data can be preserved whilst allowing many clients with various amounts of data to collaboratively train a deep neural network. A visual representation of how the original data is preserved is depicted in Figure ~\ref{fig:privacy}. Figure ~\ref{fig:privacy} (a) shows the original cifar10 image of a car. 
The original car image becomes distorted after passing through one Conv2D of the first hidden layer as shown in (b). At this stage, the outline of the car can still be recognized. However, after the data passes through the max-pooling layer, the image becomes completely deformed to the point where it can not be traced back to the original image. This is depicted in (c), where the image after the first hidden layer is hardly recognizable, therefore, preserving data privacy.

\begin{table}[t!]%

\small
\begin{center}
	\begin{tabular}{r|r}
    \toprule[1.0pt]
    Layers at end-systems & Accuracy \\
    \midrule
    Nothing (All layers are in the server) & $71.09$\,\% \\
    $L_{1}$ & $68.18$\,\% \\
    $L_{1}, L_{2}$ & $67.92$\,\% \\
    $L_{1}, L_{2}, L_{3}$ & $66.00$\,\% \\
    $L_{1}, L_{2}, L_{3}, L_{4}$ & $65.66$\,\% \\
    \bottomrule[1.0pt]
	\end{tabular}
	
\end{center}
\caption{Accuracy rate for individual layers.}
\label{accuracylayer}
\vspace{-5mm}
\end{table}

\subsection{Medical Data}\label{sec:Data}

To evaluate the proposed spatio-temporal split learning, we present the performance of split learning using three kinds of medical data, including COVID-19 chest CT scan image data, MURA image datasets, and patient’s cholesterol data. 

\begin{figure*}
    \begin{center}
    
         \subfloat[\centering COVID-19 patient CT scan]{
            \includegraphics[width=0.45\linewidth]{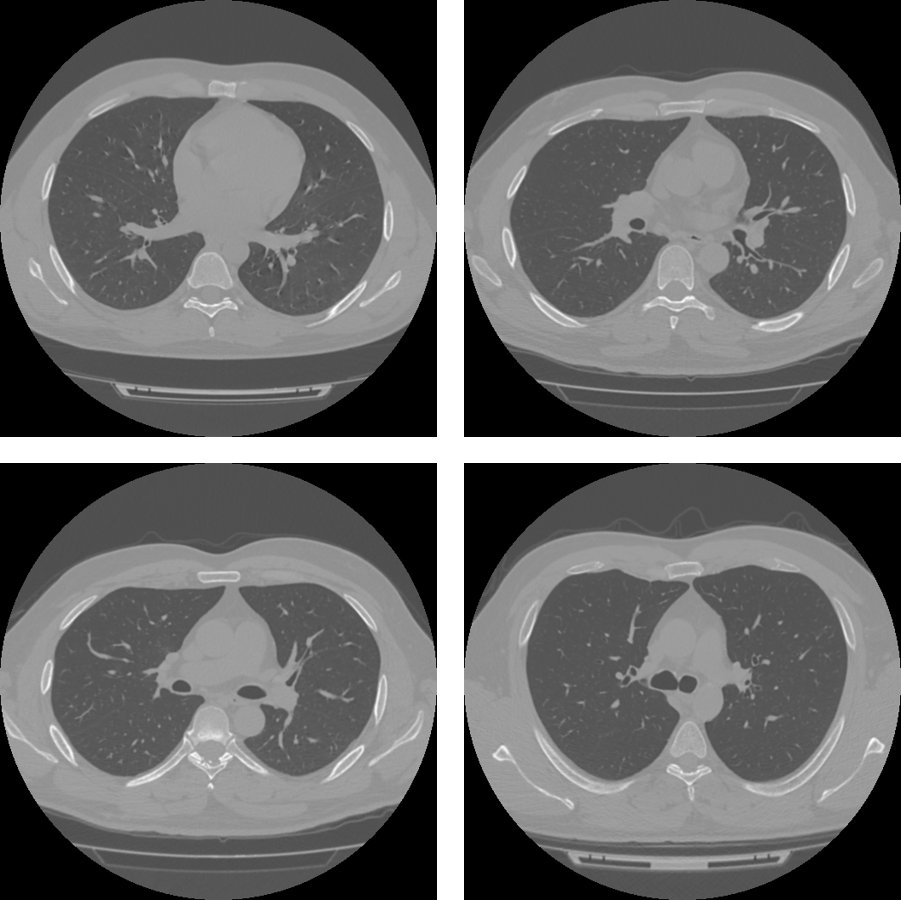}}\label{fig:covidpatient}
              \qquad
            \subfloat[\centering non-COVID-19 patient CT scan]{
            \includegraphics[width=0.45\linewidth]{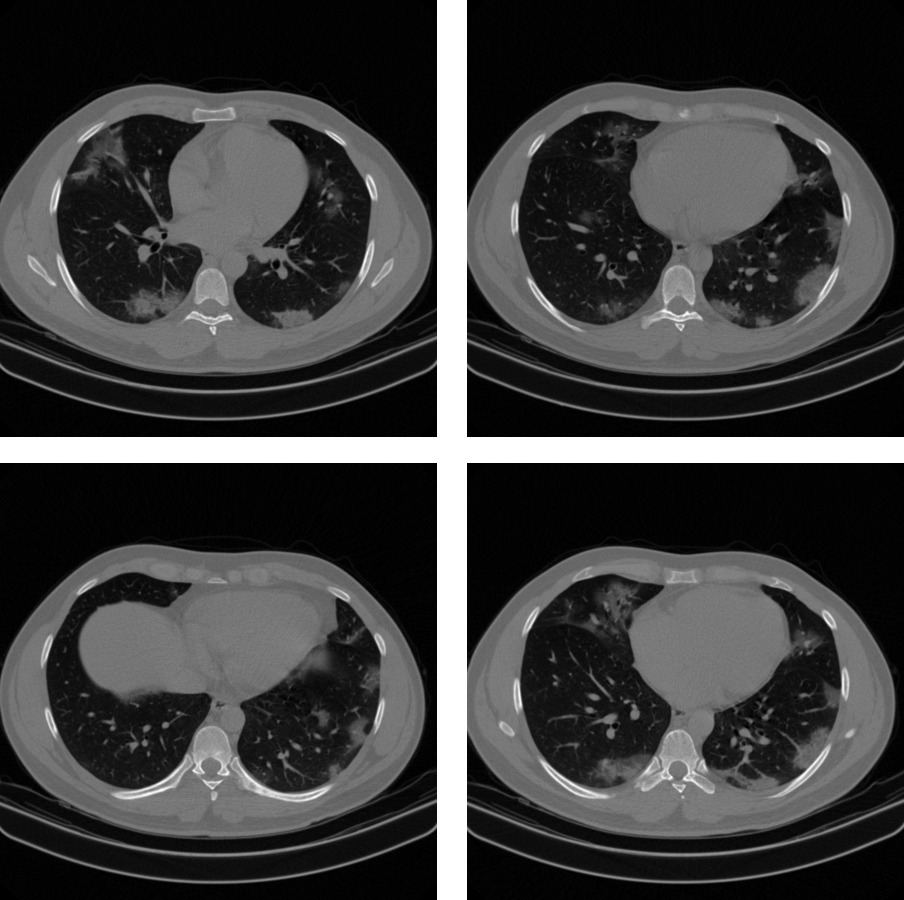}}
            \label{fig:covidpatient}
    \end{center}
    \caption{The lung CT scans of (a) COVID-19 patient and (b) non-COVID-19 patient.}
    \label{fig:covidnoncovid}
\end{figure*}

Using the COVID-19 chest CT scans, the model learns to classify the images according to lungs that are infected by the COVID-19 virus from those that are healthy, as shown in Fig. \ref{fig:covidnoncovid}. 
Those who got positive results for the COVID-19 virus, would suffer from lung complications such as pneumonia \cite{covid}. The COVID-19 chest CT scan is a dataset of lungs obtained from \cite{COVID19}. 
It is the largest COVID-19 chest CT scan that is curated from multiple public datasets~\cite{covid1, covid2, covid3, covid4, covid5, convid6}. 
This paper conducts the split learning algorithm to classify which lung image is of a patient that suffers from pneumonia contaminated by the COVID-19 and which is of a healthy patient. 
These two classes of data are used in medical research platforms to track changes in the treatment progress \cite{covid}. Hence, primarily identifying which CT image is the scan of a viral COVID-19 virus and which is of a healthy lung is key. This can be considered a binary classification problem. 
The MURA dataset is a collection of X-ray images of patient’s arms gathered by Stanford University School of Medicine and ML Group~\cite{rajpurkar2018mura}. This data provides X-rays of the elbow, finger, forearm, hand, humerus, shoulder, and wrist. The data is used to classify abnormalities in the X-ray scan. We conduct an experiment to determine whether or not the X-ray image has a crack in the bone. If the bone X-ray scan does not have a fracture, then the patient is not diagnosed with musculoskeletal disorders, if not, then the patient suffers from musculoskeletal conditions. If the patient's bone appears to have a crack, it is classified positive, as in the patient does have a musculoskeletal disorder; the X-ray image is classified as negative if the bone is in one piece. The number of MURA X-ray scans of 7 different body parts is shown in Table~\ref{tab:muratest}, where the total number, number of positive case, and negative case used in training for each body part is configured.

\begin{table} 
\begin{center}
	\centering
	\begin{tabular}{l|ccc}
    \toprule[1.0pt]
    \centering
      Category &  Total case & Positive case & Negative case\\
    \midrule[1.0pt]
    Finger  & 5,106   & 1,968  & 3,138  \\
    Hand    & 5,543   & 1,484  & 4,059  \\
    Wrist  & 9,752   & 3,987  & 5,765 \\
    Forearm & 1,825   & 661   & 1,164 \\
    Elbow    & 4,931   &2,006  & 2,925  \\
    Humerus & 1,272   &599   & 673  \\
    Shoulder&  8,379   & 4,168  &  4,211 \\
    \bottomrule[1.0pt]
	\end{tabular}
\end{center}
\caption{The number of X-ray scans done on each body part collected from MURA.}
\label{tab:muratest}
\end{table}

\begin{figure*} 
    \centering
    \includegraphics[width=0.99\linewidth]{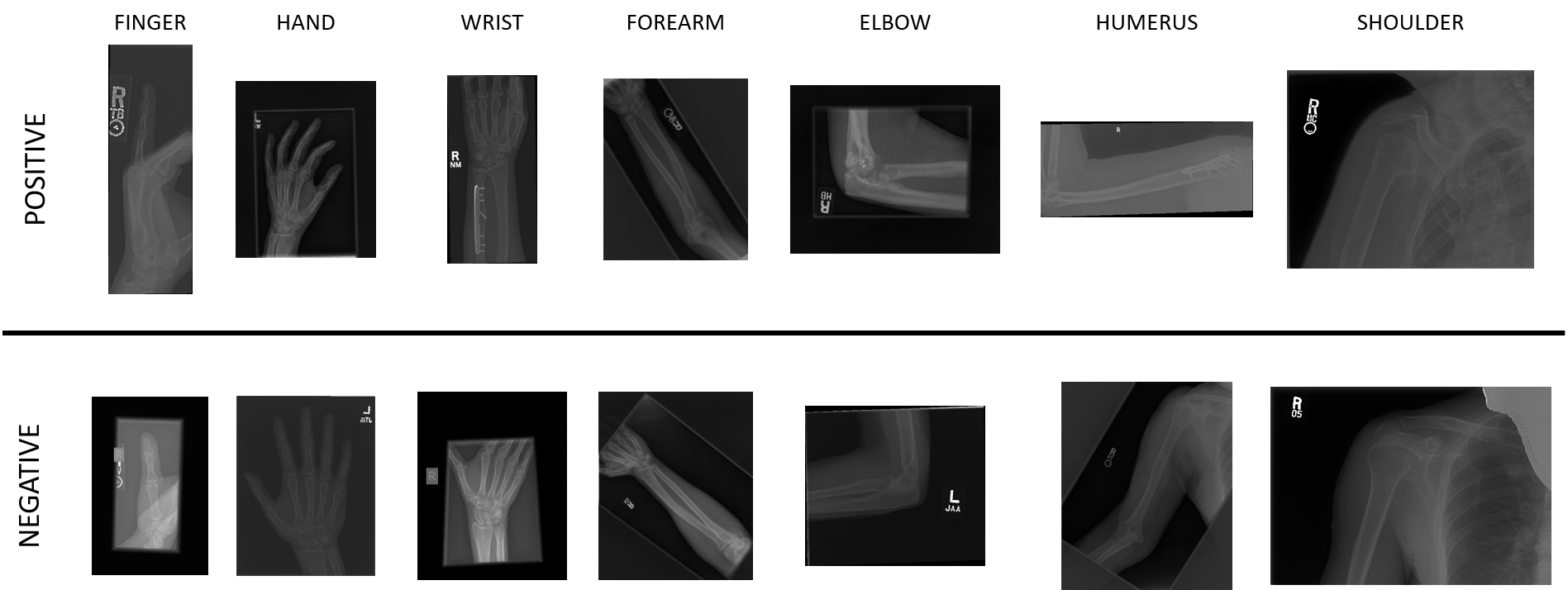}
    \caption{MURA X-ray images of the 7 body parts. The top section is positive for a fracture being present in the bones, and the bottom section is negative for an absence of a fracture in the bones.}
    \label{fig:MURA_dataset}
\end{figure*}

Before learning of a deep neural network can take place, the image data must be resized to the same size. 
COVID-19 chest CT scans and MURA X-ray data both require a pre-processing stage. 
The sets of images obtained from the open source all have different sizes, as indicated in Fig. \ref{fig:MURA_dataset}. 
In other words, the CT images vary in size and the bone X-ray images differ in size as well. 
With inputs all having different sizes, training cannot run smoothly. 
Hence, this pre-processing step is needed to reshape the images into all the same sizes. 
The images are simply resized to an appropriate size depending on the image size distribution of the original image, being careful not to loose any valuable information of the original image.

The cholesterol dataset is provided by SNUH. Note that cholesterol data is numeric, thus our learning model predicts the target cholesterol value based upon other medical information. Using the proposed spatio-temporal learning on actual patient records, we learn a general cholesterol value prediction model. 
The cholesterol dataset includes the information on the patient’s age, sex, height, weight, total cholesterol (TC), high-density lipoprotein cholesterol (HDL-C), low-density lipoprotein cholesterol (LDL-C), and triglyceride (TG). Cholesterol travels through the bloodstream using the protein called ‘lipoprotein’. LDL-C is the harmful type that collects in the walls of the blood vessels and raises the risk of heart disease and strokes \cite{heart}. Yet, not all cholesterol is lethal. The HDL-C type absorbs cholesterol and directs it to the liver where it is excreted from the body \cite{chol}.

\begin{table*}[t]
\begin{center}
	\centering
	\begin{tabular}{c|ccccccc}
    \toprule[1.0pt]
    \centering
      Index  & Age & Sex& Height & Weight & TC & HDL-C & TG \\
    \midrule[1.0pt]
    Case 1 & 62 & Male & 175.0 & 68.20 & 178 & 50 & 83  \\
    Case 2 & 80 & Male & 168.0 & 78.70 & 104 & 22 & 148  \\
    Case 3 & 56 & Male & 178.0 & 80.85 & 207 & 55 & 158  \\
    Case 4 & 73 & Female & 144.8 & 50.45 & 144 & 30 & 100  \\
    Case 5 & 66 & Male & 167.7 & 62.80 & 138 & 60 & 74   \\
    \bottomrule[1.0pt]
	\end{tabular}
\end{center}
\caption{A sample of the five original cholesterol data obtained from SNUH without the LDL-C level. Only the attributes used in the prediction model are shown.}
\label{tab:LDL-dataset}
\end{table*}

Since LDL-C is the dangerous type of cholesterol, medical infrastructures want to predict the level of LDL-C in the blood using the attributes such as age, sex, height, weight, TC, TG, and HDL-C. There are various forms of equations that calculate LDL-C value using the values of TC, HDL-C and TG. However, these equations are only valid under certain conditions \cite{LDLequation}. The experiment shows that the LDL-C level estimated using the prediction model with the same attributes, provides a similar value to the original LDL-C value. This numerical data also requires pre-processing. In the pre-processing step, we selectively choose the attributes for predicting the LDL-C level. Five examples from the pre-processed cholesterol data are shown in Table~\ref{tab:LDL-dataset}. Note that only the attributes used to predict the LDL-C level are presented.

\subsection{Experiment Setup}

The experiment conducted in this paper compares the performance of our proposed spatio-temporal split learning algorithm to the conventional split learning algorithm using three different datasets--COVID-19 chest CT scans, MURA bone X-ray scans, and cholesterol data.

\subsubsection{Overcoming data-imbalance}
To simulate an environment where several medical organizations collaborate to build a high-accuracy medical AI model, the experiment assumes three clients connected to one centralized server. This basic setup applies to all three datasets considered in this paper. Each dataset is divided in a 7:2:1 ratio. In other words, out of the three clients, one hospital will possess 70~\% of the data, the other hospital will possess 20~\% of the data, and the remaining one hospital will possess only 10~\% of the data. Note that this data is divided into this ratio after 10~\% of the data is saved as the validation set to tune the model and 10~\% of the data is saved as the test dataset to evaluate the performance of our spatio-temporal split learning. Hence, 80~\% of the remaining data is then divided into this 7:2:1 ratio as the inputs to the client. In a single-client split learning method, the hospital with 10~\% of data is bound to be overfitted. Yet, even if one hospital only has 10~\% of data, overfitting does not become a problem with our learning method as hospitals with 70~\% and 20~\% of the dataset jointly train the deep neural network placed in the server. This data division is applied equally to the COVID-19 CT scans, MURA X-ray scans, and cholesterol data. 

\subsubsection{Model configuration for image data} 
For CNN to classify images, we conduct pre-processing to reshape all the images to the same sizes. COVID-19 chest CT scans with a variety of image sizes are all re-sized to $64\times64\times1$. The MURA bone X-rays are pre-processed to a size of $224\times224\times1$.

\begin{table*}[t]
\begin{center}
	\centering
	\begin{tabular}{l|ccc}
    \toprule[1.0pt]
    \centering
       Parameters & Covid-19 chest & MURA & Cholesterol\\
    \midrule[1.0pt]
    Epochs  & 100 & 50 & 200 \\
     Loss & Binary crossentropy & Binary crossentropy & MSE  \\
    Activation function & Sigmoid & Sigmoid & Leaky ReLU \\
     Batch size & 64 & 128 & 2,048 \\
    Input Size & 64 $\times$ 64 $\times$ 1 & 224 $\times$ 224 $\times$ 1  & 326,032  \\
    Model & Custom classification & VGG19~\cite{simonyan2014very} & Custom regression  \\
    \bottomrule[1.0pt]
	\end{tabular}
\end{center}
\caption{The deep neural network setup for each COVID-19, MURA and cholesterol data.}
\label{tab:setup}
\end{table*}

After the pre-processing step of scaling, the neural network is established. The COVID-19 chest CT scans are trained using a custom model with a total of 5 convolutional layers. In using our spatio-temporal split learning algorithm, the client will train up to the first layer, and the server will continue training along 4 of the convolutional layers. This CNN architecture is setup up using the following specifications. The input image shape is set to a size of $64\times64\times1$. The number of epochs, which is the number of times the algorithm trains on the dataset, is set to 100. For the loss function, binary crossentropy function is selected, which is a widely used loss function for a classification problem. The sigmoid function is used as the activation function in this classification model. The purpose of the activation function is to introduce non-linear data in the training process. The batch size, which is the number of training examples utilized in one iteration, is set to 64. The parameter values set for our experimental model is summarized in Table~\ref{tab:setup}.

A similar approach is undertaken with the MURA dataset. The collection of X-ray images of the patient’s arms is fed into the deep neural network to identify a presence of a fracture in the bones. The X-ray image is classified as positive if there is a crack or broken bone, and negative if the patient’s arm is intact and in one piece. This X-ray data is learned using VGG19, which is comparably deeper in its neural net structure than the previous CNN classification architectures. VGG19 is a variant of VGGNet, which significantly reduces the number of parameters in the convolutional layers to improve convergence speed. Hence, it has a total of 17 convolutional layers, where the majority of the computations are conducted within the 16 layers placed at the servers end. The number of epochs is 50, hence the model completes 50 complete passes of the X-ray scans throughout the algorithm. Since the VGG19 is also a classification model, the loss and activation functions are the same as the COVID-19 classification setup. The shape of the original input image is set to $224\times224\times1$. Again, the architecture setting is summarized in Table~\ref{tab:setup}. 
As the images for both COVID-19 and X-ray scans undergo the deep neural network for classification, the images reduce in size. The max-pooling layer used in both the CNN models, reduces the image size in half. Hence, a feature map of size $32\times32\times1$ for the COVID-19 CT scan and a feature map of size $112\times112\times1$  for the X-ray scan is transferred to the centralized server, where the rest of the convolutional layers are computed for classification.

\subsubsection{Model configuration for numerical data} 

The model trained using the cholesterol data is a predictive model. Based on the information of age, sex, height, weight, TC, TG, and HDL-C of a patient, the level of LDL-C in the bloodstream can be predicted. The learning of the prediction model is conducted with 200 epochs. The MSE is selected as the loss function, the LeakyReLU function is used as the activation function in this regression model. The batch size, which is the number of training examples utilized in one iteration, is set to 2,048. 
Mean square logarithmic error (MSLE) is generally used for the loss function as an indication of whether or not the prediction model is training well. A loss function is an evaluation index of how adequately the model makes predictions based on the training dataset. It helps to make informed decisions when tuning the algorithm as it numerically shows how much the predicted value differs from the actual value. If the algorithm models the dataset well, then the value of the loss function is a lower number; if the model is poor at making predictions, the loss function outputs a higher value. MSLE is calculated by taking the squared difference between the log-transformed true and predicted values. The equation for MSLE is expressed below:

\begin{equation}
    MSLE = \frac{1}{N} \sum_{i=0}^{N} (\log(y_i + 1) - \log(\hat{y}_i + 1))^{2}
\label{eq:msle} 
\end{equation}
where $y_i$ and $\hat{y}_i$ are the true and predicted values of LDL-C, respectively.

We use several other metrics to compare our spatio-temporal split learning algorithm to the baseline, single-client split learning. We use root mean squared logarithmic error (RMSLE) as another means to compare the loss values between our model and the baseline. Due to the nature of logarithms, the RMSLE calculates the relative error between the actual LDL-C value and the predicted LDL-C value. Since it only measures the relative error, the magnitude of the error value is not considered, making it more robust to outliers. 
Other similar loss functions such as RMSE, spikes in value when met with an outlier, making it more susceptible to record large error values. 

With this property, the RMSLE incurs a greater penalty when the predicted value is lower than the actual value; and incurs a lower penalty when the predicted value is greater than the actual value. This is great in our case where the LDL-C level, which is a harmful cholesterol in the body, is predicted. Medical centers would be better off predicting a larger level of LDL-C level rather than a lower LDL-C level and go through an extra screening process instead of merely dismissing the patient as healthy if the prediction value is lower than the actual. For precautions, predicting a higher level of a harmful substance is safer since the results can be inspected again. Therefore, RMSLE is suitable for evaluating our model for cholesterol data, and it is formulated as follows.

\begin{equation}
    RMSLE = \sqrt{\frac{1}{N} \sum_{i=0}^{N} (\log(y_i + 1) - \log(\hat{y}_i + 1))^{2}}.
\end{equation}

Symmetric mean absolute percentage error (sMAPE) is another tool to measure the difference between the true and predicted value \cite{kim2020multiscale}. This is also a measure of accuracy based on percentage or relative errors. It is calculated by the formula:

\begin{equation}
    sMAPE = \frac{100\%}{n} \sum_{i=1}^{N} \bigg ( \frac  {|y_i - \hat{y}_i|}{|y_i| + |\hat{y}_i|} \bigg ).
\end{equation}

The performance of our spaito-temporal split learning is compared to the single-client split learning, where only one client is involved in the split learning process with one centralized server. Just as image and numerical data use different models to train their data, image data is assessed quite differently from numerical data. The evaluation criteria of experiments for these two different split learning algorithms are the loss rates and percentage accuracy for the image data. Cholesterol level data is evaluated by comparing loss functions MSLE, RMSLE and sMAPE. 

Table~\ref{tab:setup} summarizes the experiment setting such as the number of epochs, batch size, shape size, type of loss function, and activation function used to train the model with the three datasets. These same experimental preconditions are applied to evaluate the performance of the single-client split learning.

\subsection{Experiment Results} 
The experimental results using cholesterol levels, COVID-19 chest CT scan, and MURA dataset are presented in this section. 

\subsubsection{Classification accuracy}

\begin{figure}[t]
     \centering
     \begin{subfigure}[b]{0.49\textwidth}
         \centering
         \includegraphics[width=\textwidth]{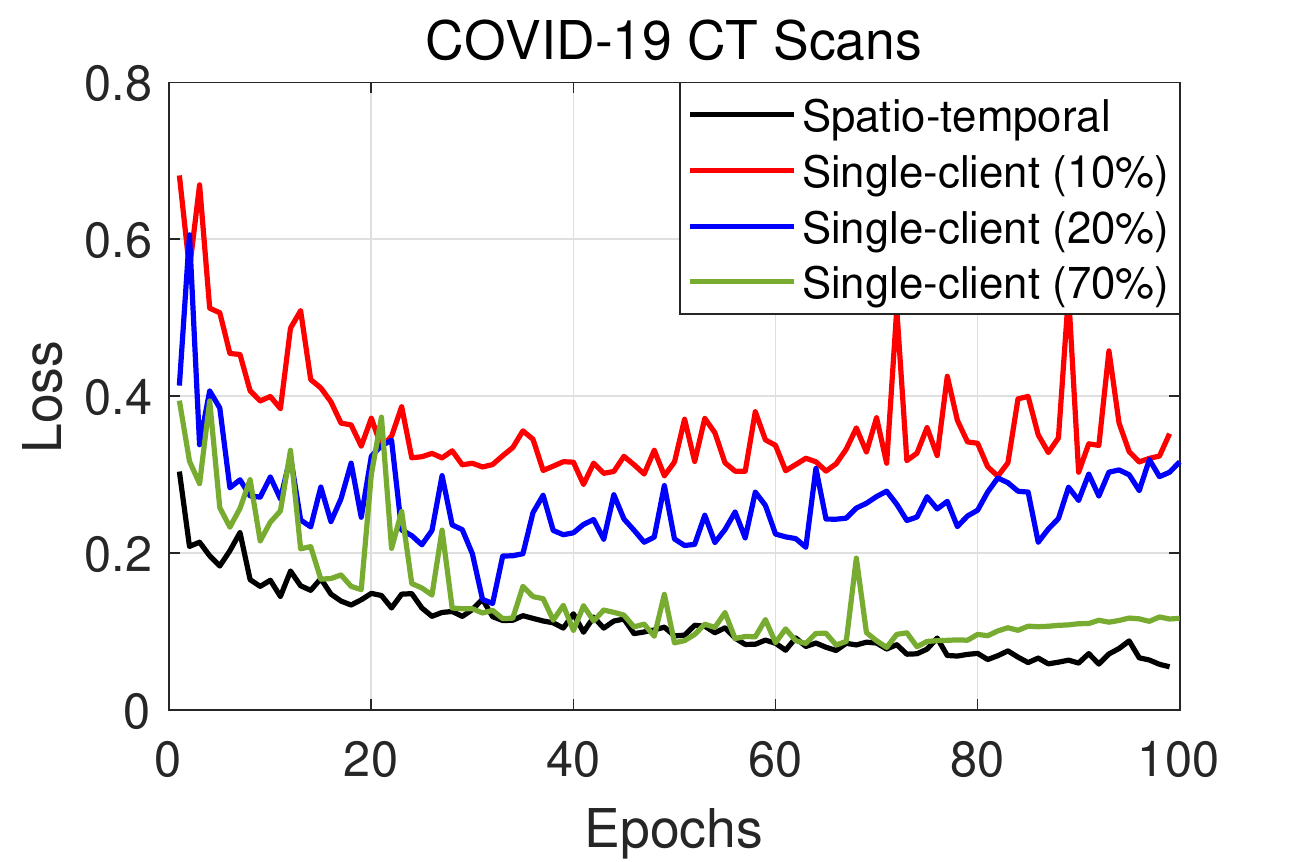}
         \caption{Loss}
         \label{fig:covidresultloss}
     \end{subfigure}
     \hfill
     \begin{subfigure}[b]{0.49\textwidth}
         \centering
         \includegraphics[width=\textwidth]{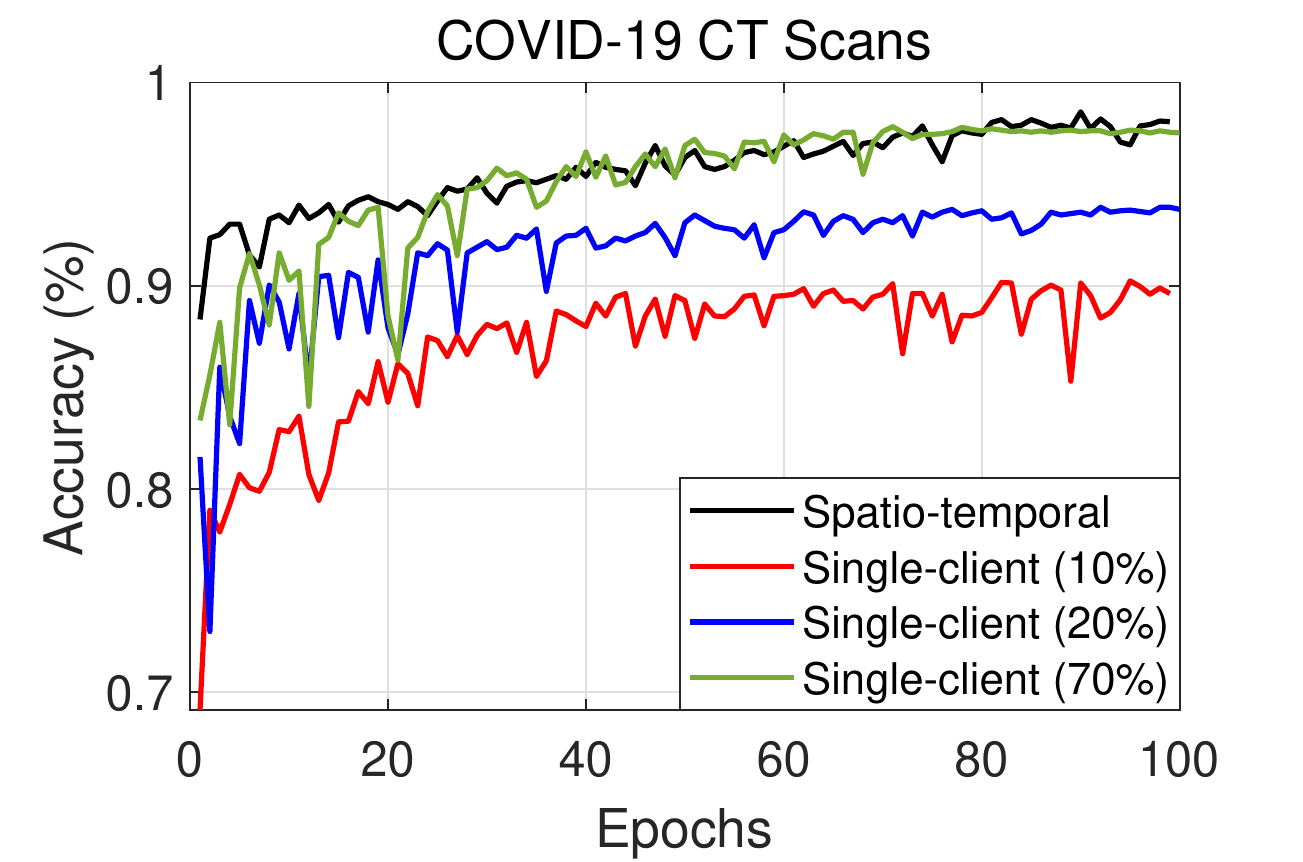}
        
         \caption{Accuracy}
         \label{fig:covidresultacc}
     \end{subfigure}
     \hfill
        \caption{Graph comparing (a) loss value between single-client and spatio-temporal split learning; and (b) accuracy value between single-client and spatio-temporal split learning for CNN model training using COVID-19 CT scans.}
        \label{fig:covidresultcap}
\end{figure}
Graphs comparing the loss and accuracy of the two split learning algorithms using COVID-19 chest CT scans are shown in Fig.~\ref{fig:covidresultcap}. The graphs show the performance of all the discussed cases. Our proposed multi-client spatio-temporal split learning algorithm, which is represented by the black curve, is compared to the single clients with 10\%, 20\%, and 70\% of data, which are shown in red, blue, and green, respectively. As can be seen in Fig.~\ref{fig:covidresultcap} (a), the loss value for all the single-client models fluctuates quite aggressively over the course of epochs. There is an obvious decrease in loss with our model, where the loss value itself is lower and it converges at a steady rate compared to the baseline model. It can also be observed that after a certain amount of epochs, the loss values for the three single-client models, whether it be the 10\%, 20\%, or 70\% data, start to elevate, whereas our proposed method continues to decrease.This is due to the insufficient amount of data available for training the single-client split learning model, which leads to overfitting. Furthermore, the red curve, which is a depiction of a single-client model with the least amount of data--10\%--has the highest loss value. Comparing this with the black curve clearly demonstrates the effect of data-imbalance.

The accuracy in classifying whether a patient tests positive to COVID-19 just by analyzing chest CT scans is depicted in Fig.~\ref{fig:covidresultcap} (b). The classification accuracy of our proposed model is the highest. It also has steady learning progress than the single-client models, which spikes quite a bit even after 80 epochs. As repeatedly stated through this paper, multi-site spatio-temporal split learning resolves the issue of data-imbalance. Comparing the black curve with the red curve very distinctly shows the effect of data-imbalance. Fig.~\ref{fig:covidresultcap} (b) shows that the classification accuracy of a single-client split learning increases as the amount of data increases from 10\% or 20\% or 70\%, with 70\% of data being almost high as our proposed method. Hence, even with multiple clients participating in training the deep neural network model, the classification accuracy is high at around 98.5\% with the advantage of data security. 

To highlight the exceptional performance of our multi-client spatio-temporal split learning algorithm, this paper compares the experiment results simulated using FL. As meticulously detailed in Section~\ref{sec:related}, FL encompasses a completely different learning process from split learning. An experiment is conducted to analyze the classification accuracy using COVID-19 chest CT scans with FL. To make a fair and unbiased comparison, the experimental setup for FL is identical to the setup of our proposed method used for classifying whether or not a patient is diagnosed with COVID-19 after analyzing their CT scan. With the exact same setup to our learning method, FL achieved a classification accuracy of only 95.7\%, whilst split learning obtained an accuracy up to 98.5\%. This is 2.8\% lower than the accuracy achieved by our multi-client spatio-temporal split learning process. The results are summarized in Table~\ref{tab:fl}. Therefore, a greater performance is accomplished when utilizing split learning, especially multi-client spatio-temporal kind of spit learning, whilst allowing computationally-limited hospitals such as small private hospitals to achieve a near-perfect classification.

The single-client split learning algorithm considers the case where the one client might not hold an abundance of data. Hence, in this experiment, the single-client is set to possess only 10\% or 20\% or 70\% of the data. In other words, our spatio-temporal split learning algorithm with the three clients having a 7:2:1 data division is tested against the single-client split learning that has one-tenth, one-twentieth, or one-seventieth of the data. This is strategically done to demonstrate the reality of hospitals having data-imbalance. Since the hospital with only 10\% of the data, compared to hospitals that have 20\%  and 70\% of the data, possess very small amounts of training data, the overfitting of the model occurs. Hence, the red curve has a higher loss than our black curve. The same reasoning is applied to the accuracy rate as well. The red, data-insufficient hospital, curve depicts a lower classification accuracy since it does not have enough data to make accurate groupings of COVID-19 chest CT scans. On the other hand, our model in black, which exhibits a higher accuracy percentage than hospitals with 10\% or 20\% or 70\% of data all collaborate together to train one large CNN model in the server. Whilst all the appropriate accuracy levels are met, the privacy of these medical images is also protected. 
 
\begin{table} 
\begin{center}
	\centering
	\begin{tabular}{l|ccc}
    \toprule[1.0pt]
    \centering
      &  Federated Learning & Split Learning\\
    \midrule[1.0pt]
    Accuracy (\%)   & 95.7   & 98.5   \\
    Setup    & \multicolumn{2}{c}{Refer to Table~\ref{tab:setup}} \\ 
    \bottomrule[1.0pt] 
	\end{tabular}
\end{center}
\caption{A comparison of classification accuracy between federated learning and split learning.}
\label{tab:fl}
\end{table} 

\begin{figure} 
    \centering
        \includegraphics[width=1.0\columnwidth]{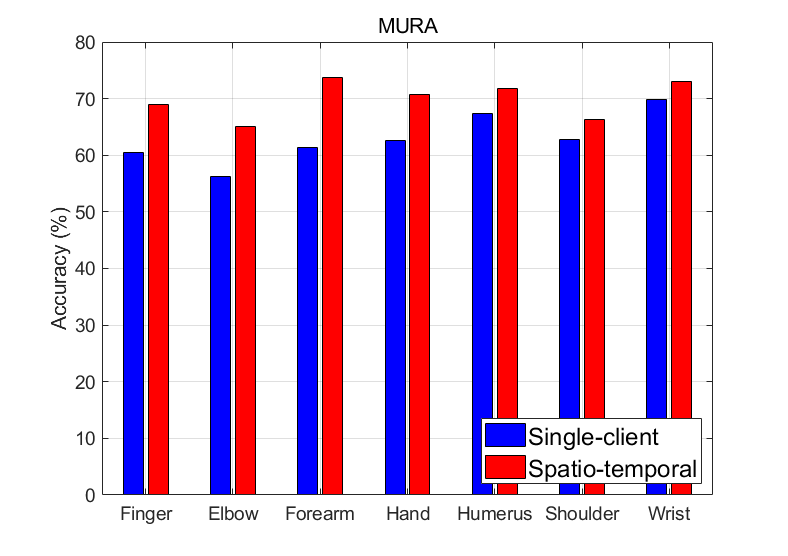}

    \caption{A bar graph of accuracy level comparing single-client and spatio-temporal split learning for MURA X-ray scans. The x-axis labels each 7 of the body parts.}
    \label{Fig:muraaccbar}
\end{figure}

MURA data consists of X-ray scans of 7 body parts--finger, hand, wrist, forearm, elbow, humerus, and shoulder--that we all trained. The accuracy rate of single-client and spatio-temporal split learning is graphed in Fig.~\ref{Fig:muraaccbar} for each part. Just like the analysis made with the COVID-19 CT scans, our algorithm shows a superior accuracy rate than that of the single-client split learning algorithm. The blue bars are of the single-client with again, only having 10\% of the data and the yellow illustrates our algorithm. Since ours is tested against a hospital with a significantly smaller number of data available, which leads to overfitting, the accuracy rate of spatio-temporal split learning is higher for every body part.

\begin{table*} [t]
\begin{center}
	\centering
	\begin{tabular}{cc|ccccccc}
    \toprule[1.0pt]
    \centering
       && Finger & Elbow & Forearm & Hand & Humerus & Shoulder & Wrist  \\
    \midrule[1.0pt]
    \multirow{2}{*}{Accuracy (\%)} & Single-client  & 60.5& 56.3 & 61.4 & 62.6 & 67.3 & 62.8 & 69.9\\
    & Spatio-temporal  & 68.9 & 65.1 & 73.7 & 70.8 & 71.8 & 66.4 & 73.1\\
    \bottomrule[1.0pt]
	\end{tabular}
\end{center}
\caption{Accuracy data points for training MURA dataset with single-client and spatio-temporal split learning.}
\label{tab:muraacc}
\end{table*}
Table~\ref{tab:muraacc} shows the specific accuracy value in classifying the MURA bone X-ray images into two groups: fractured and unfractured bones. The single-client split learning appears to have a slightly lower accuracy level than the accuracy level of our proposed system. In identifying a crack in the finger, the single-client split learning has an accuracy of 60.5\%  and our multi-client split learning system achieves an accuracy of 68.9\%. Our model performs better by 8.4\%, which is a  significant gap in performance in classifying medical data. Looking at the performance of classifying fractures with an elbow X-ray image, the single-client split learning has a 56.3\% accuracy rate  whereas our algorithm has a 65.1\% accuracy rate. Once again, there is an increase of 8.8\% in classification performance in our multi-client split learning. As repeatedly stated throughout this paper, the increase in performance goes hand in hand with the protection of personal information. Without sacrificing classification accuracy, the same preservation of training data cannot be achieved.

\subsubsection{Preserving privacy of medical data} 
It has repeatedly been stated throughout this paper that the privacy of the original data is preserved. This section visually shows how the original medical image gets distorted to the degree it becomes hardly discernible, not even being able to recognize the outline of a lung or bone let alone determine the presence of a crack.

\begin{figure} 
\center
    \subfloat[\centering Original COVID-19 image]{
    \includegraphics[width=0.4\linewidth]{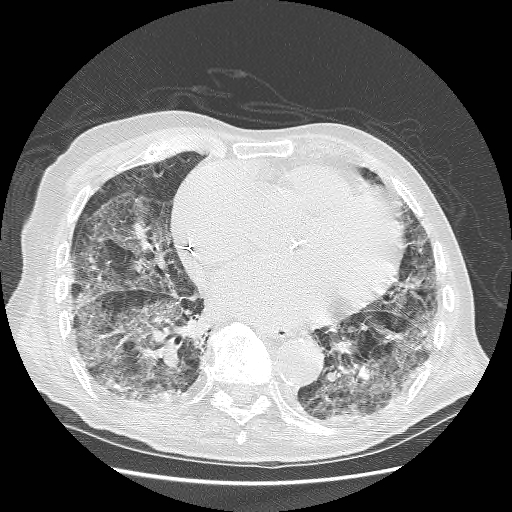}
    }
        \qquad
    \subfloat[\centering Distorted COVID-19 image]{
    \includegraphics[width=0.4\linewidth]{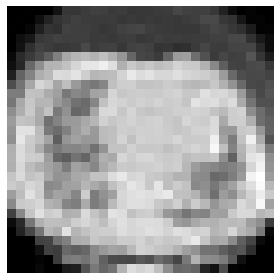}
    }
    \caption{ The (a) original CT image of a COVID-19 patient and (b) COVID-19 chest CT image that is processed by the first hidden layer at the client.}
    \label{img:covidct}
\end{figure}

\begin{figure} 
\center
    \subfloat[\centering Original elbow scan]{
        \includegraphics[width=0.45\linewidth]{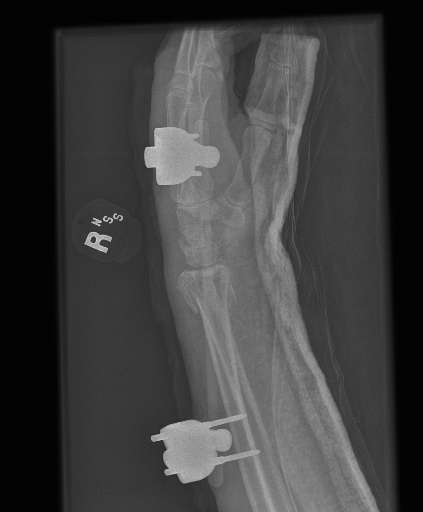}
    }
    \subfloat[\centering Distorted elbow image]{
        \includegraphics[width=0.45\linewidth]{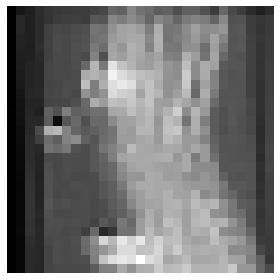}
    }
    \caption{ The (a) original image of a negatively grouped X-ray scan of an elbow and (b) that image after passing through the first hidden layer in the client side.}
    \label{img:wrist}
\end{figure}

The original image of the chest CT scan of a COVID-19 patient and one that has passed through the first and only hidden layer at the client side is shown in Fig.~\ref{img:covidct}. Fig.~\ref{img:covidct} (a) shows the original CT scan of a patient with COVID-19 and (b) shows an image of lungs after it goes through the first hidden layer located at the client, hence it is the image that is transferred to the server. It is apparent that the second image is hardly recognizable. Therefore, as the server continues to train the CNN model, these medical images become even more distorted. The original scans are not shared among hospitals, and only image (b) from each client is sent to the server, hence, preserving the privacy of sensitive medical information. The same phenomenon can be observed in Fig.~\ref{img:wrist}. Fig.~\ref{img:wrist} (b) is the image that is sent to the server, which is unrecognizable. Thus, the original medical data, Fig.~\ref{img:wrist} (a), is protected during split learning. Note that medical images usually have high-resolution because we need high-accuracy in diagnosis.

\subsubsection{Prediction accuracy} 

To check the performance of the regression model learned with single-client and spatio-temporal split learning method trained by the numerical cholesterol data, the MSLE is calculated. Since MSLE is a loss function that calculates the relative error between predicted and ground truth values, a lower MSLE value refers to a better, more accurate model. The red represents the performance of our model and the blue data points are the single-client split learning algorithm. The value of the final MSLE, RMSLE, and sMAPE are shown in Table~\ref{tab:cholmetrics}. 

\begin{figure*} 
    \centering
    \subfloat[\centering CDF for MSLE Loss]{
        \includegraphics[width=0.676\columnwidth]{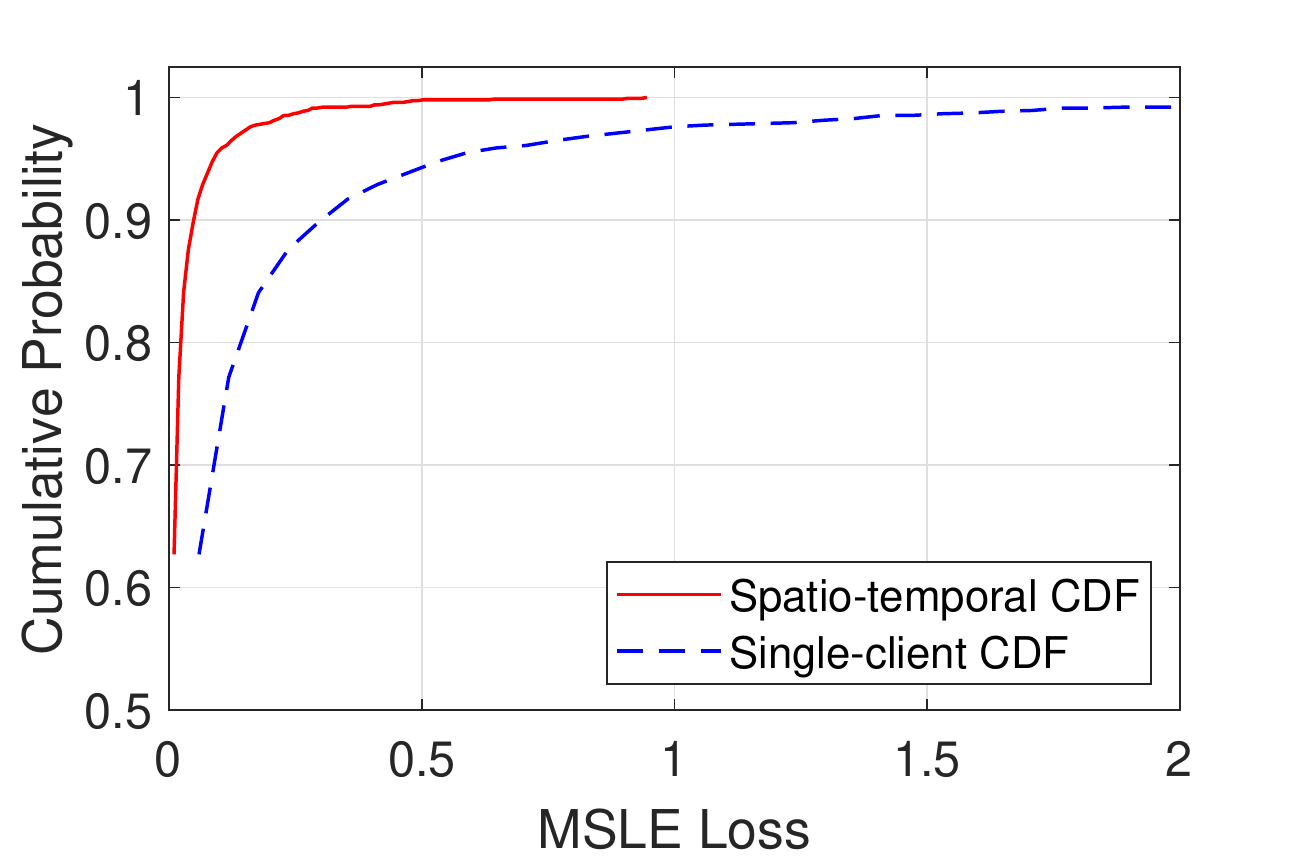}
    }
    \centering
    \subfloat[\centering CDF for RMSLE Loss]{
        \includegraphics[width=0.676\columnwidth]{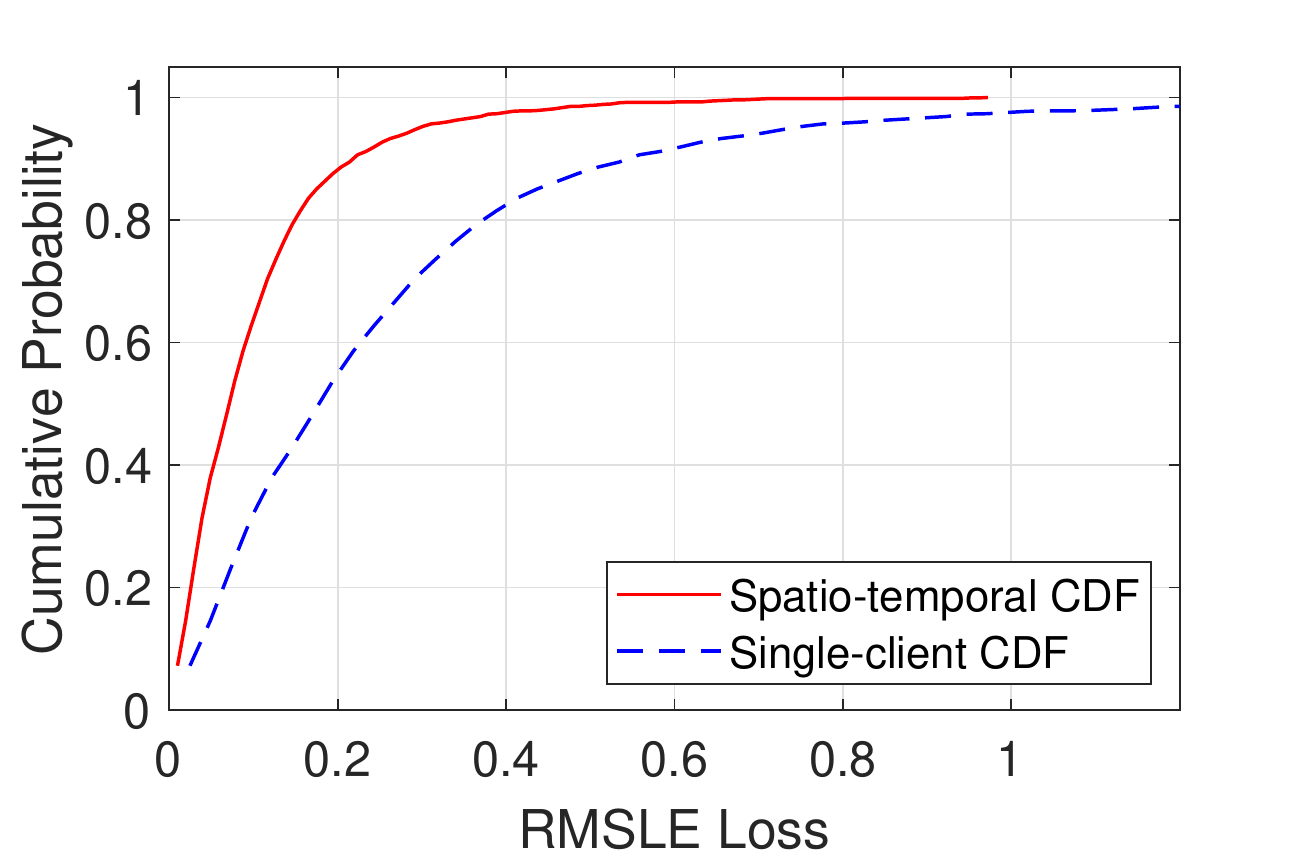}
    }
    \centering
    \subfloat[\centering CDF for sMAPE]{
        \includegraphics[width=0.676\columnwidth]{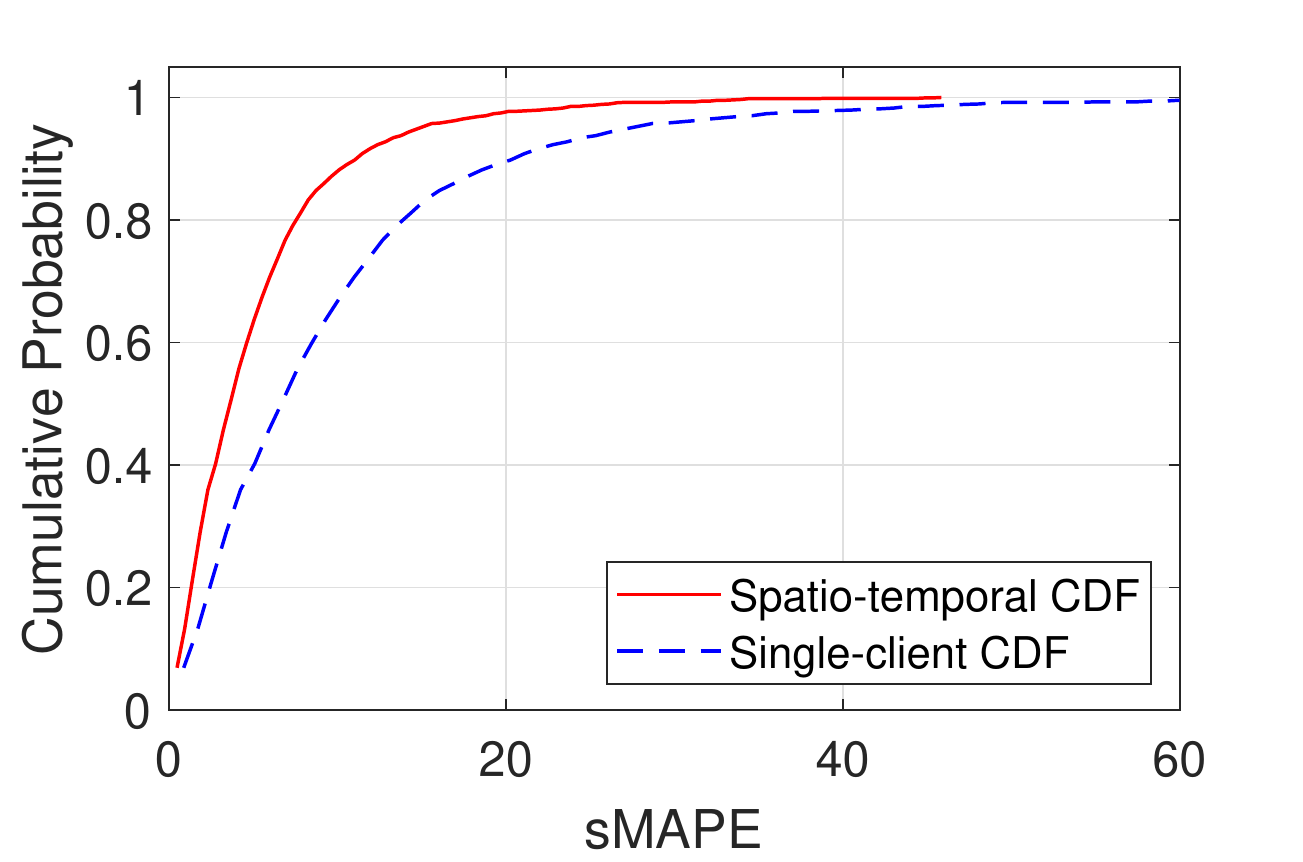}
    }
    \caption{CDF graph comparing (a) MSLE values (b) RMSLE values and (c) sMAPE of single-client and spatio-temporal split learning for regression model using cholesterol data.}
    \label{Fig:cholcdf}
\end{figure*}

The graph in Fig.~\ref{Fig:cholcdf} shows the cumulative distribution function (CDF) of the losses found using MSLE and RMSLE for the cholesterol data. This graph represents the distribution of the loss values obtained after testing with 1508 data. Having a steep gradient at the beginning of the curve means that there is a large distribution of loss with a lower value, which would mean that the model has predicted the cholesterol levels close to the target value. As seen in Fig.~\ref{Fig:cholcdf} (a), the CDF graph of MSLE loss function shows that the blue curve representing the spatio-temporal learning process has a much steeper gradient than the red single-client model. The same pattern can be observed in Fig.~\ref{Fig:cholcdf} (b), where it can be interpreted as our model predicting the LDL-C level better than the single-client model, thus giving us a higher probability of the loss being around a very small value of between 0 and 0.2 for the RMSLE loss values. Fig.~\ref{Fig:cholcdf} (c) represents the cumulative probability distribution of the relative error values calculated for single-client and spatio-temporal split learning.

\begin{figure*} 
    \centering
    \subfloat[\centering PDF for MSLE Loss]{
        \includegraphics[width=0.676\columnwidth]{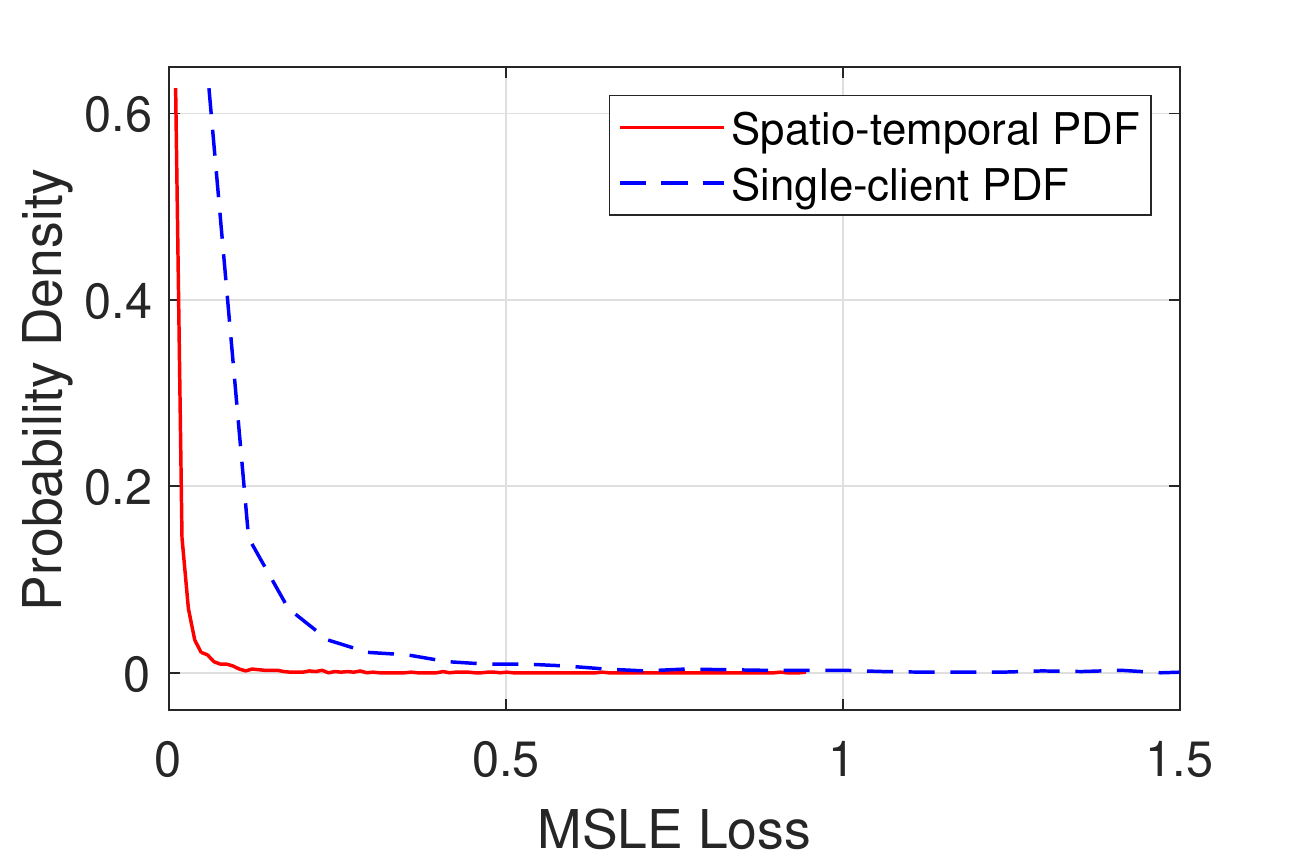}
    }
    \centering
    \subfloat[\centering PDF for RMSLE Loss]{
        \includegraphics[width=0.676\columnwidth]{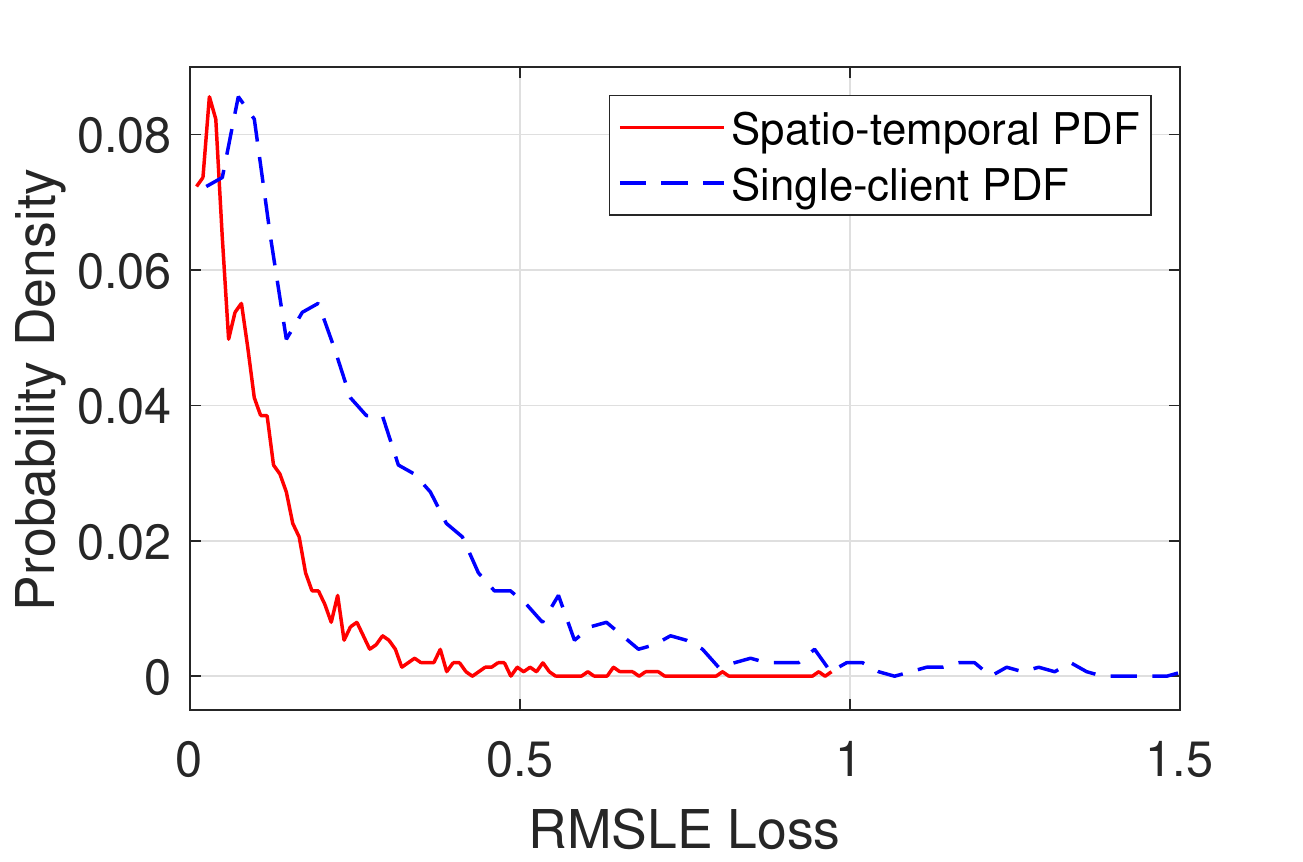}
    }
    \centering
    \subfloat[\centering PDF for sMAPE]{
        \includegraphics[width=0.676\columnwidth]{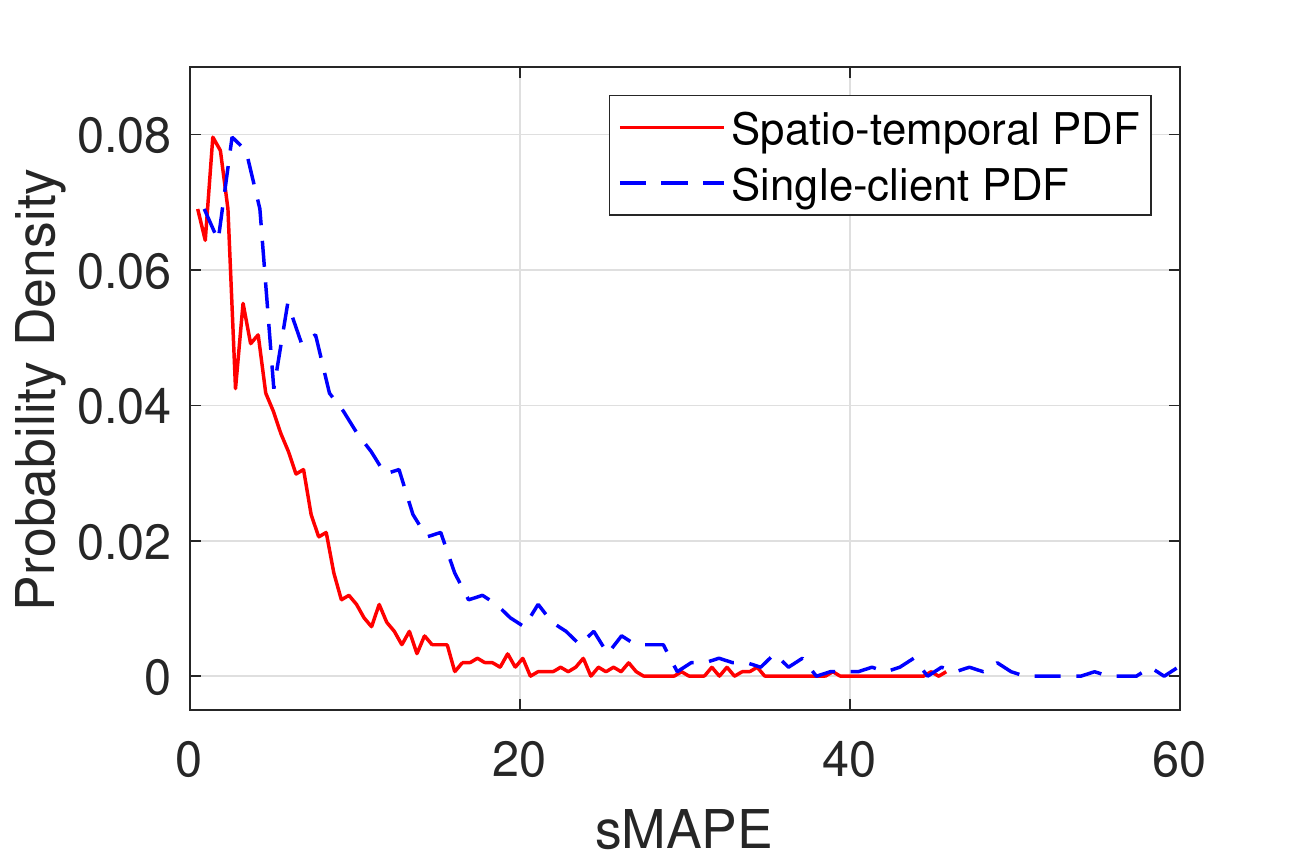}
    }
    \caption{PDF graph comparing (a) MSLE values (b) RMSLE values and (c) sMAPE of single-client and spatio-temporal split learning for regression model using cholesterol data.}
    \label{Fig:cholpdf}
\end{figure*}

A similar comparison between our model and the single-client model is made using Probability Distribution Functions (PDF) graphs using MSLE and RMSLE as depicted in Fig.~\ref{Fig:cholpdf}. This is a graph that shows the probability distribution of MSLE loss and RMSLE loss. Therefore, in Fig.~\ref{Fig:cholpdf} (a), our model drawn in blue demonstrates there is a greater possibility of the MSLE loss value being smaller than the baseline model. The PDF graph exhibited using the RMSLE loss function in Fig.~\ref{Fig:cholpdf} (b) is analogous to that of the previous explanations. Graph (c) also shows that our model has a higher probability near the lower error value, which proves our model is better at predicting the LDL-C level in the blood.  

The spatio-temporal split learning model has a lower loss for all three metrics, MSLE, RMSLE and sMAPE, compared to the single-client model as summarized in Table~\ref{tab:cholmetrics}. This phenomenon is once again due to the fact that the single-client has a significantly lower number of data, which leads to overfitting. Since the single-client only trains the model using its small amount of medical information, it is only fit to take in those types of data. Hence, when other data from the test set are inputted to predict the LDL-C value, it shows a considerable amount of loss as depicted in Fig.~\ref{Fig:cholcdf} and Fig.~\ref{Fig:cholpdf}.

\begin{table} 
\begin{center}
	\centering
	\begin{tabular}{l|ccc}
    \toprule[1.0pt]
    \centering
       Metrics & MSLE & RMSLE & sMAPE (\%) \\
    \midrule[1.0pt]
    Single-client  & 0.1428 & 0.3062 & 15.1403 \\
    Spatio-temporal & 0.0216 & 0.1002 &  5.0609 \\
    \bottomrule[1.0pt]
	\end{tabular}
\end{center}
\caption{Comparison of single-client and spatio-temporal split learning loss values using MSLE, RMSLE, and sMAPE loss functions.}
\label{tab:cholmetrics}
\end{table}

The results for training the cholesterol level are analyzed below. Table~\ref{tab:muratest} shows five patient’s original cholesterol data, with the LDL-C level omitted since that is the value we want to predict, obtained from SNUH. Once the client runs its one and only hidden layer, the gradient or one-dimensional feature map is transferred to the server. Table~\ref{tab:muratest} includes several personal information such as age, sex, height, and weight. However, the information that is exposed to the public, which is the gradient update, is an encrypted data that does not reveal or hint out anything about the original data. Hence, even if malicious attacks target the medical data, the hackers will only be able to hack the information that the client sends to the server. The cyber attackers will not be able to trace the information back to the original data that contains sensitive personal information using the parameter updates in our split learning setting. Privacy preservation is firmly secured.

\section{Conclusion and Discussion}\label{sec:5}

This paper introduces an innovative learning method that trains a deep neural network without fear of exposing original raw data. Unlike regular deep learning where the client learns a single deep neural network, split learning shares the learning step between the clients and a centralized server. Our spatio-temporal split learning algorithm helps multiple clients collaboratively train a deep neural network with a centralized server. In doing so, the raw data in each client is never shared with each other and only the encrypted feature maps are exposed to the network, thus, complete and utter protection of original data is achieved. This approach is especially promising in machine learning with privacy, where protection of user information is critical such as medical or finance applications.

In addition, our spatio-temporal split learning study shows that it efficiently alleviates an overfitting problem. Since multiple clients can participate in model training, clients can collaborate to build much higher-accuracy models with much light-weight processing. Further, the proposed split learning works with clients with imbalance data volume. Finally, the proposed split learning is versatile--the split learning is general to learn image or numeric data, and the user can wisely choose the neural network model according to the target application. With actual medical data, such as patient’s COVID-19 chest CT scans, X-ray bone scans, and cholesterol levels, we present how the proposed split learning achieves high-accuracy machine learning with privacy. 
Note that our algorithm is applicable in any setting whether medical, finance, banks, or insurance--where protection of personal records is crucial, data-imbalance between participating clients is an issue and diverse types of data must be processed. 

For future studies, we hope to find the effects of varying the number of clients participating in split learning. This can be extended to include studies on the performance changes seen according to variations made to the data ratio of training datasets. Furthermore, we hope to explore the implications of utilizing differential privacy, a metric that measures the degree of privacy protected in future studies.

\bibliographystyle{IEEEtran}
\bibliography{ref_ML}

\begin{IEEEbiography}[{\includegraphics[width=1in,height=1.25in,clip,keepaspectratio]{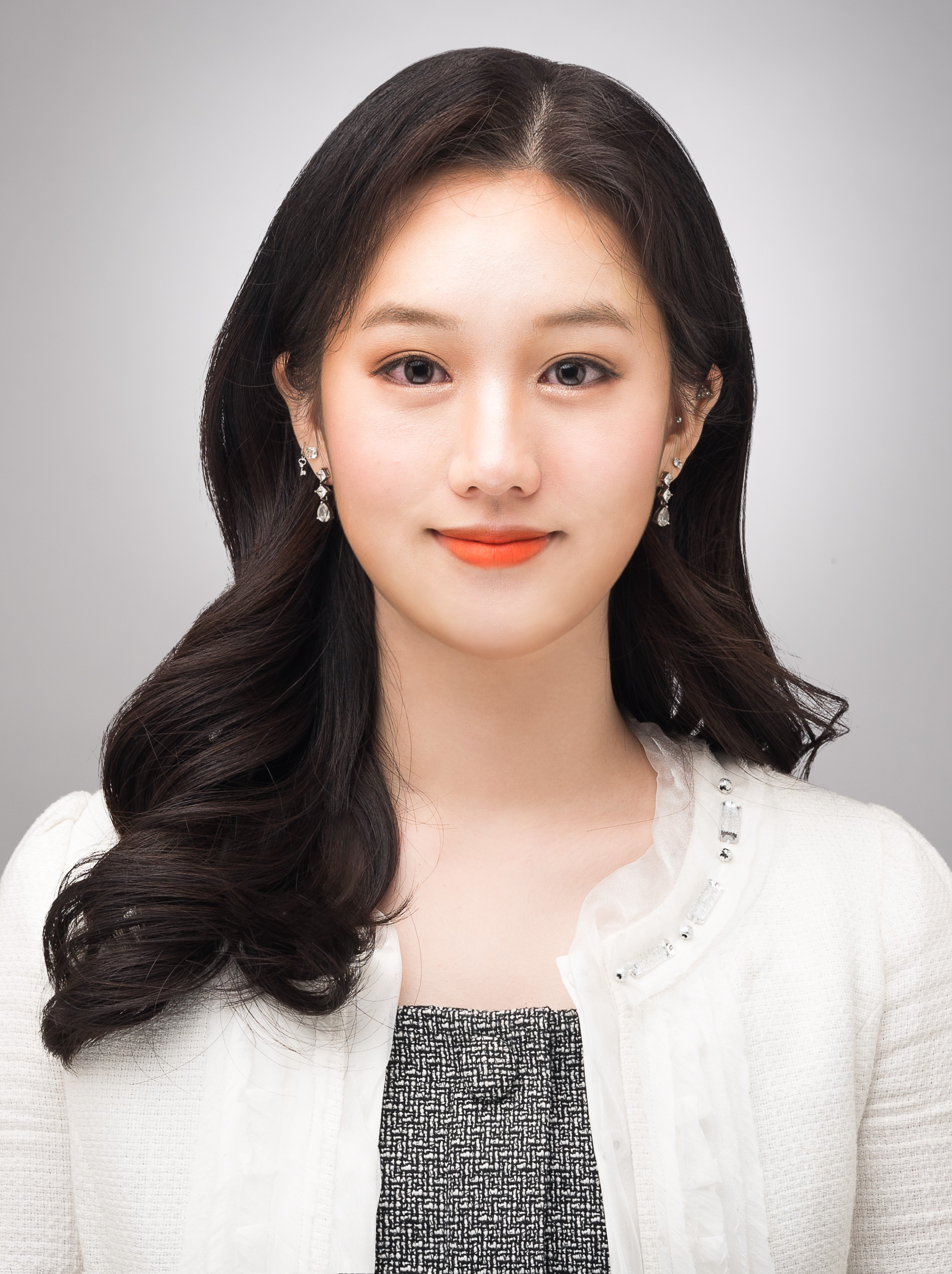}}]{Yoo Jeong Ha} has been with Korea University, Seoul, South Korea, since March 2017, and she is currently a Ph.D. candidate at the Department of Electrical and Computer Engineering, Seoul, South Korea. She received a B.S. degree in Mechanical Engineering, College of Engineering from Korea University, Seoul, South Korea, in 2020. Her research focuses on and is not limited to computer vision, deep learning algorithms, and their applications in mobility and communications. 

\end{IEEEbiography}

\begin{IEEEbiography}[{\includegraphics[width=1in,height=1.25in,clip,keepaspectratio]{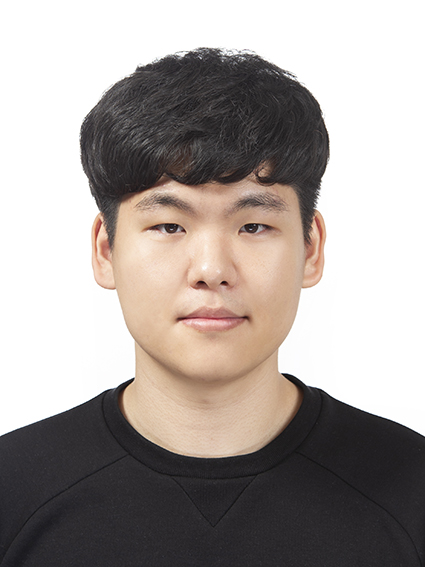}}]{Minjae Yoo} is currently an M.S. student at the School of Electrical Engineering with Korea University, Seoul, South Korea. He received a B.S. degree in Electrical Engineering, Kyungpook National University, Daegu, South Korea.  His research focuses on and is not limited to computer vision, deep learning algorithms, and their applications in mobility and communications. 
\end{IEEEbiography}

\begin{IEEEbiography}[{\includegraphics[width=1in,height=1.25in,clip,keepaspectratio]{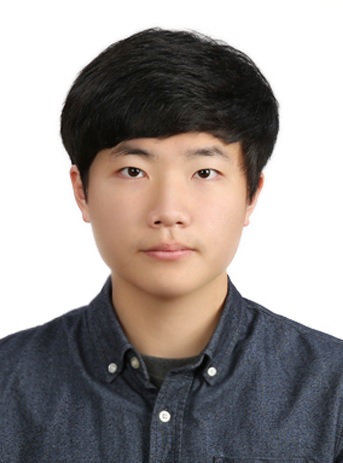}}]{GuSang Lee} is currently an undergraduate research intern at the Artificial Intelligence and Mobility (AIM) Laboratory at Korea University, Seoul, South Korea. He will be receiving his B.S. degree in Electrical Engineering, Korea University, Seoul, South Korea. His research interests include deep learning, AR/VR, metaverse and its applications. 
\end{IEEEbiography}

\begin{IEEEbiography}[{\includegraphics[width=1in,height=1.25in,clip,keepaspectratio]{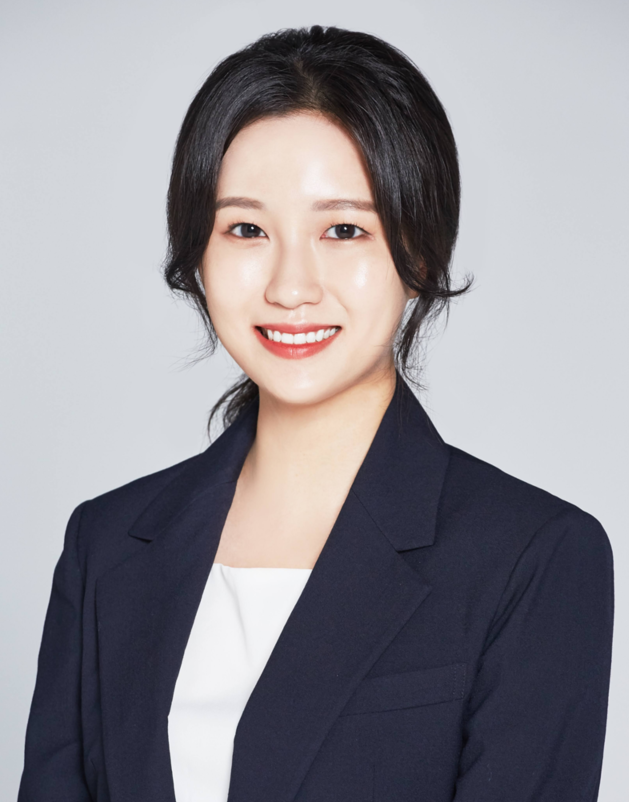}}]{Soyi Jung} has been an assistant professor at the School of Software, Hallym University, Chuncheon, Republic of Korea, since September 2021. She also holds a visiting scholar position at Donald Bren School of Information and Computer Sciences, University of California, Irvine, CA, USA, from 2021 to 2022. She was a research professor at Korea University, Seoul, Republic of Korea, during 2021. She was also a researcher at Korea Testing and Research (KTR) Institute, Gwacheon, Republic of Korea, from 2015 to 2016. 
She received her B.S., M.S., and Ph.D. degrees in electrical and computer engineering from Ajou University, Suwon, Republic of Korea, in 2013, 2015, and 2021, respectively. 

Her current research interests include network optimization for autonomous vehicles communications, distributed system analysis, big-data processing platforms, and probabilistic access analysis. She was a recipient of Best Paper Award by KICS (2015), Young Women Researcher Award by WISET and KICS (2015), Bronze Paper Award from IEEE Seoul Section Student Paper Contest (2018), ICT Paper Contest Award by Electronic Times (2019), and IEEE ICOIN Best Paper Award (2021).
\end{IEEEbiography}

\begin{IEEEbiography}[{\includegraphics[width=1in,height=1.25in,clip,keepaspectratio]{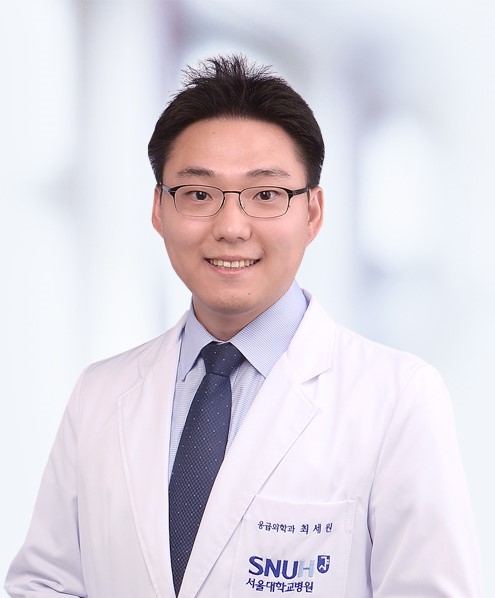}}]{Sae Won Choi}, M.D., is an Assistant Professor at Seoul National University Hospital's Office of Hospital Information. He works on the implementation of the hospital's new IT initiatives and the maintenance and management of the hospital information system. His research interest is in the implementation of artificial intelligence and machine learning to hospital systems. He is a board certified emergency medicine physician, completing his training at Seoul National University Hospital. 
\end{IEEEbiography}

\begin{IEEEbiography}[{\includegraphics[width=1in,height=1.25in,clip,keepaspectratio]{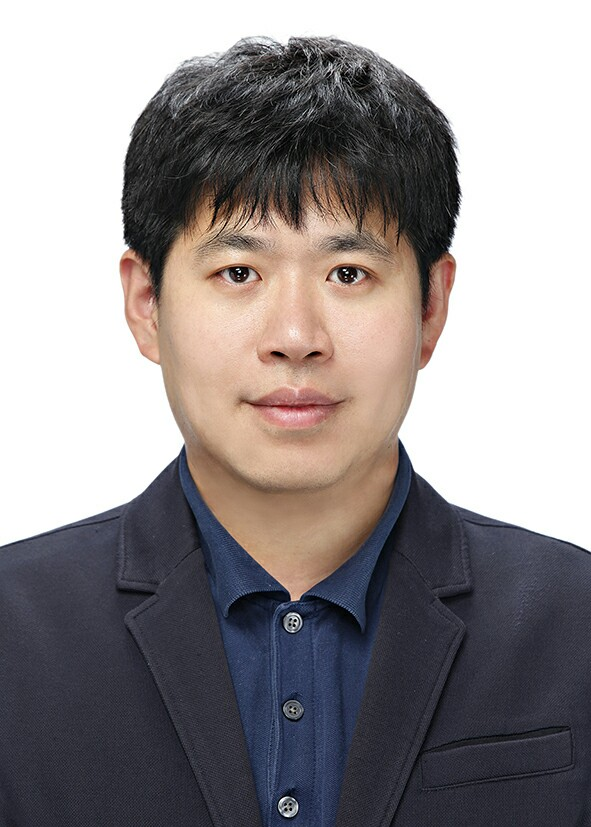}}]{Joongheon Kim} (M'06--SM'18) has been with Korea University, Seoul, Korea, since 2019, and he is currently an associate professor at the Department of Electrical and Computer Engineering. He is also a vice director of Artificial Intelligence Engineering Research Center and a dean of Center for Teaching and Learning at Korea University, Seoul, Korea. 
He received the B.S. and M.S. degrees in Computer Science and Engineering from Korea University, Seoul, Korea, in 2004 and 2006, respectively; and the Ph.D. degree in Computer Science from the University of Southern California (USC), Los Angeles, CA, USA, in 2014. 
Before joining Korea University, he was with LG Electronics (Seoul, Korea, 2006--2009), InterDigital (San Diego, CA, USA, 2012), Intel Corporation (Santa Clara in Silicon Valley, CA, USA, 2013--2016), and Chung-Ang University (Seoul, Korea, 2016--2019). 

He is a senior member of the IEEE, and serves as an associate editor for \textit{IEEE Transactions on Vehicular Technology} and a guest editor for \textit{IEEE Communications Standards Magazine}. He internationally published more than 90 journals, 120 conference papers, and 6 book chapters. He also holds more than 50 patents, majorly for 60\,GHz millimeter-wave IEEE 802.11ad and IEEE 802.11ay standardization. 

He was a recipient of Annenberg Graduate Fellowship with his Ph.D. admission from USC (2009), 
Intel Corporation Next Generation and Standards (NGS) Division Recognition Award (2015), Haedong Young Scholar Award by KICS (The Korean Institute of Communications and Information Sciences) (2018), IEEE Vehicular Technology Society (VTS) Seoul Chapter Award (2019), Outstanding Contribution Award by KICS (2019), Paper Awards from IEEE Seoul Section Student Paper Contests (2019, 2020), Best Teaching Awards by Korea University (Top 5\% in Fall-2019, Top 20\% in Fall-2020), \textit{IEEE Systems Journal} Best Paper Award (2020), IEEE ICOIN Best Paper Award (2021), and Haedong Paper Award by KICS (2021).
\end{IEEEbiography}

\begin{IEEEbiography}[{\includegraphics[width=1in,height=1.25in,clip]{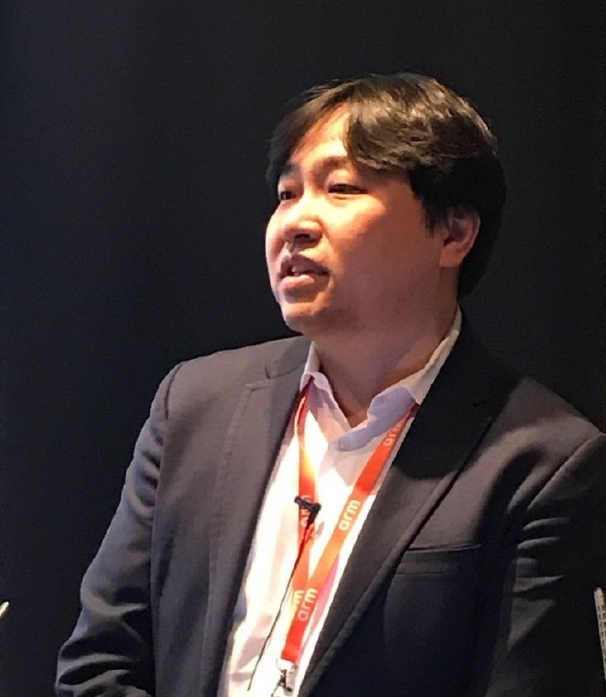}}]{Seehwan Yoo} received his B.S., M.S., and Ph.D. degrees in computer science from Korea University, Seoul, Republic of Korea, in 2002, 2004, and 2013, respectively. 
From 2013 to 2014, he worked for the Software Platform Laboratory at LG Electronics, Seoul, Republic of Korea.
He is currently an associate professor in the Department of Mobile Systems Engineering, Dankook University, Yong-in, Republic of Korea.
His current research interests include cloud computing systems and system security for AI/ML applications and services. He actively involves in open source community works, such as secure boot, Xen, KVM/Qemu. 
\end{IEEEbiography}

\EOD

\end{document}